\documentclass{article}
\usepackage{arxiv}
\usepackage{amsmath,amsfonts,bm}

\def\eqref#1{equation~\ref{#1}}
\def\1{\bm{1}}

\def\vs{{\bm{s}}}

\DeclareMathAlphabet{\mathsfit}{\encodingdefault}{\sfdefault}{m}{sl}
\SetMathAlphabet{\mathsfit}{bold}{\encodingdefault}{\sfdefault}{bx}{n}

\pdfoutput=1

\usepackage{microtype}
\usepackage{graphicx}
\usepackage{subcaption}
\usepackage{booktabs} % for professional tables

\usepackage{wrapfig}
\usepackage{xspace}
\usepackage{colortbl}
\usepackage{amssymb}
\usepackage{algpseudocode}
\usepackage{arydshln}
\usepackage{tabularx}
\usepackage{rotating} 
\usepackage{caption}

\usepackage{xcolor}
\definecolor{darkblue}{rgb}{0, 0.12, 0.55}
\definecolor{darkgreen}{rgb}{0, 0.55, 0.12}
\definecolor{darkred}{rgb}{0.6,0,0}
\definecolor{darkgreen}{rgb}{0,0.6,0}
\definecolor{Gray}{gray}{0.9}
\definecolor{my_green}{rgb}{0,0.6,0}
\definecolor{my_gray}{gray}{0.9} 
\usepackage[breaklinks=true,
            colorlinks,
            linkcolor = darkred,
            urlcolor  = darkblue, 
            citecolor = teal,
            bookmarks = false]{hyperref}
\usepackage[numbers]{natbib}
\usepackage{url}

\usepackage{adjustbox}
\usepackage{multirow}
\usepackage{array}
\usepackage{bm}
\usepackage{bbm}
\usepackage{cleveref}
\usepackage{amsthm,amsmath,amssymb}
\usepackage{algorithm}
\usepackage{algorithmicx}
\usepackage{algpseudocode}

\usepackage{subcaption}
\usepackage{colortbl}
\usepackage{adjustbox}
\usepackage{makecell}
\usepackage{pifont}
\usepackage{tcolorbox}
\usepackage{setspace}
\usepackage{fancybox}
\usepackage{tocloft}
\usepackage{enumitem}

\makeatletter
\DeclareRobustCommand\onedot{\futurelet\@let@token\@onedot}
\def\@onedot{\ifx\@let@token.\else.\null\fi\xspace}
\def\eg{\emph{e.g}\onedot}

\def\etc{\emph{etc}\onedot} \def\vs{\emph{vs}\onedot}
 
\def\etal{\emph{et al}\onedot}
\makeatother

\newlength{\mysize}

\allowdisplaybreaks

\theoremstyle{definition}

\DeclareMathOperator*{\argmax}{arg\,max}

\title{ITS3D: Inference-Time Scaling for Text-Guided \\ 3D Diffusion Models}

\author{
Zhenglin Zhou$^{1}$ \ \ \ \
Fan Ma$^{1}$ \ \ \ \
Xiaobo Xia$^{2}$ \ \ \ \
Hehe Fan$^{1}$ \ \ \ \
Yi Yang$^{1\dagger}$ \ \ \ \
Tat-Seng Chua$^{2}$ \\\
$^1$ReLER, Zhejiang University \quad\quad
$^2$National University of Singapore \\
\tt\small \{zhenglinzhou, hehefan, yangyics\}@zju.edu.cn \quad \{xbx, dcscts\}@nus.edu.sg \quad flowerfan524@gmail.com
}

\begin{document}

\maketitle
\def\thefootnote{$\dagger$}\footnotetext{Corresponding author.}

\begin{abstract}

We explore inference-time scaling in text-guided 3D diffusion models to enhance generative quality without additional training.
To this end, we introduce \textbf{ITS3D}, a framework that formulates the task as an optimization problem to identify the most effective Gaussian noise input. 
The framework is driven by a verifier-guided search algorithm, where the search algorithm iteratively refines noise candidates based on verifier feedback.
To address the inherent challenges of 3D generation, we introduce three techniques for improved stability, efficiency, and exploration capability.
1) Gaussian normalization is applied to stabilize the search process.
It corrects distribution shifts when noise candidates deviate from a standard Gaussian distribution during iterative updates.
2) The high-dimensional nature of the 3D search space increases computational complexity. To mitigate this, a singular value decomposition-based compression technique is employed to reduce dimensionality while preserving effective search directions. 
3) To further prevent convergence to suboptimal local minima, a singular space reset mechanism dynamically updates the search space based on diversity measures.
Extensive experiments demonstrate that ITS3D enhances text-to-3D generation quality, which shows the potential of computationally efficient search methods in generative processes. 
The source code is available at \url{https://github.com/ZhenglinZhou/ITS3D}.

\end{abstract}

\section{Introduction}
\label{sec:intro}

Generative models have revolutionized numerous fields, \eg, computer vision~\cite{kingma2013auto,rezende2015variational,goodfellow2020generative,chen2020generative,xia2022pluralistic,li2022out,wang2025lavin} and natural language processing~\cite{radford2019language,brown2020gpt3,achiam2023gpt,luo2025next,liu2025mtp}, by learning to generate high-quality samples from underlying data distributions. 
Among these, diffusion models~\cite{ho2020denoising,song2020denoising,ho2022classifier,chen2023humanmac} have emerged as a dominant paradigm for generating continuous data, such as audio~\cite{kong2020diffwave,liu2023audioldm}, image~\cite{ho2022imagen,stable_diffusion}, video~\cite{ho2022video,blattmann2023stable,lyu2024etau,lyu2025met}, and 3D assets~\cite{luo2021diffusion,zhang2024gaussiancube}. 
Their success can be largely attributed to the ability to scale effectively during training, by leveraging large datasets, increased computational resources, and growing model sizes~\cite{henighan2020scaling,peebles2023scalable,liu2025principled,liu2025continual,zhang2024clay}.

By leveraging large-scale 3D datasets~\cite{deitke2023objaverse,deitke2024objaverse}, 3D diffusion models~\cite{luo2021diffusion,zhou20213d,nichol2022point,wang2022rodin,jun2023shap,shue20233d,ntavelis2023autodecoding} enable users to generate high-quality 3D assets within minutes, which play an important role in applications such as AR/VR, gaming, and 3D printing.
Notably, when the generated results are unsatisfactory, a common but inefficient solution is to repeatedly change the random seed until achieving a desirable outcome. 
However, this trial-and-error approach lacks efficiency and scalability. 

\begin{figure}[t]
    \centering
    \vspace{-1.0em}
    \includegraphics[width=1.0\linewidth]{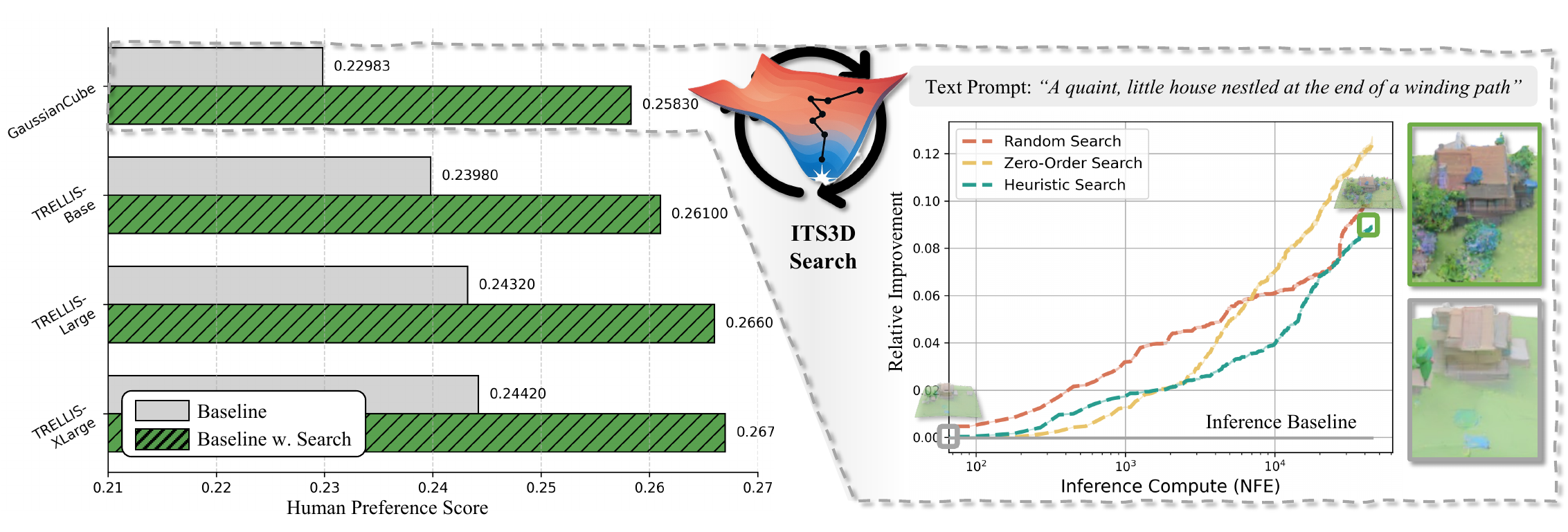}
    \vspace{-1.0em}
    \caption{
    Inspired by the achievements of inference-time scaling in LLMs~\cite{snell2024scaling} and 2D diffusion~\cite{ma2025inference,guo2025can}, we explore its application in 3D diffusion models (\eg, GaussianCube~\cite{zhang2024gaussiancube} and TRELLIS~\cite{xiang2024structured}). 
    We evaluate random search, zero-order search, and heuristic search~\cite{yang2009firefly} on GPTEval3D~\cite{wu2024gpt}, demonstrating consistent improvements in human preference scores~\cite{wu2023human}. 
    Our search framework achieves quality enhancement across all settings without requiring additional training.
    }
    \vspace{-1em}
    \label{fig:teaser}
\end{figure}

Fortunately, recent studies suggest that inference-time scaling~\cite{snell2024scaling}, allocating additional compute resources during inference, can further enhance the output quality. Building on this, we explore inference-time scaling in 3D diffusion models, optimizing both the efficiency and quality of 3D generation through structured search. 
Note that while well-studied in 2D diffusion models~\cite{ma2025inference,guo2025can}, its potential in 3D remains largely unexplored. To bridge this gap, we develop an inference-time scaling framework for text-guided 3D diffusion models, incorporating two key components: a search verifier and a search algorithm. The search algorithm iteratively refines candidate noise based on feedback from the search verifier, which assesses the quality of the generated 3D asset. As demonstrated in Figure~\ref{fig:teaser}, an inference-time scaling law emerges in 3D diffusion models, where increasing the number of search iterations consistently leads to improved generation results.

Particularly, to mitigate distribution shifts caused by noise candidates deviating from a normal Gaussian distribution in iterative search updates, we introduce a Gaussian normalization operation. It keeps the input noise in a standard normal distribution, stabilizing the search process and improving generation quality. In addition, we investigate the role of search space in inference-time scaling. Compared to image generation, 3D generation operates in a higher-dimensional search space~\cite{zhang2024gaussiancube,xiang2024structured}, which makes brute-force search computationally impractical. To tackle this challenge, we employ singular value decomposition~(SVD) to reduce the dimensionality of the search space. By optimizing only the singular values while keeping singular vectors fixed, we further improve the generation fidelity. Furthermore, to prevent iterative search algorithms from getting stuck in suboptimal local minima, we introduce a singular space reset mechanism. This mechanism updates the singular search space by selecting the best candidate when the diversity, measured by the variance of the verifier score, among search candidates decreases. By promoting continuous exploration, it ultimately enhances scaling performance.

We evaluate our framework using GPTEval3D~\cite{wu2024gpt}, which is a benchmark dataset containing diverse textual prompts for 3D generation. To assess generation performance, we leverage multiple evaluation metrics, including CLIPScore~\cite{radford2021clip} for image-text alignment, ImageReward~\cite{xu2024imagereward} for human preference modeling, and GPT4V Grader~\cite{wu2024gpt} for a holistic evaluation of text, geometry, and texture coherence. Our framework achieves state-of-the-art performance across these metrics, outperforming both optimization-based and feed-forward baselines.
By leveraging structured search and scaling techniques, we provide a practical and efficient strategy to improve 3D generation models, paving the way for future research on inference-time optimization in generative modeling. 
The contributions of this paper are as follows.
% Before delving into details, we clearly emphasize our contributions as follows. 
\begin{itemize}[leftmargin=2.5em]
    \item Conceptually, we present a systematic exploration of search-based inference-time scaling in 3D diffusion models for text-to-3D generation, which offers new insights into optimizing the inference process.
    \item Technically, we propose an innovative inference-time scaling framework for text-to-3D generation. It integrates Gaussian normalization, search space compression, and space reset, which collectively enhance the effectiveness of inference-time scaling.
    \item  Empirically, we conduct extensive experiments to analyze the inference-time scaling in text-to-3D generation. Our results highlight the superior performance of our framework over state-of-the-art baselines across multiple evaluation metrics. Comprehensive ablation studies and discussions on key design choices are also provided.
\end{itemize}

\section{Related Work}
\label{sec:related_work}

\noindent\textbf{Text-to-3D Generation.}
Early in development, due to the challenges of collecting large-scale 3D data, researchers focus on lifting 2D supervision to construct 3D assets.
Building on prior knowledge from text-to-2D models~\cite{stable_diffusion}, several methods~\cite{wang2023prolificdreamer, yu2023text, zhu2023hifa, katzir2023noise, chung2023luciddreamer, wu2024consistent3d, zhou2025dreamdpo}, such as DreamField~\cite{jain2021dreamfields}, DreamFusion~\cite{poole2022dreamfusion}, and SJC~\cite{wang2022sjc}, have been proposed to generate 3D objects guided by text prompts~\cite{li20233dsurvey}.
More recently, with the availability of large-scale 3D datasets~\cite{deitke2023objaverse, deitke2024objaverse}, multi-view diffusion models~\cite{liu2023zero, shi2023mvdream, liu2024one, liu2023syncdreamer, long2024wonder3d} have further improved the efficiency of 3D generation.
Building on these advancements, large reconstruction models~\cite{hong2023lrm, li2023instant3d, zou2024triplane, tang2024lgm, wang2024crm, xu2024grm} have been developed to reconstruct 3D assets from multi-view inputs, which significantly enhances generation efficiency.
In addition to these reconstruction-based methods, another line of work explores training 3D diffusion models. Some methods first utilize variational auto-encoders (VAEs) to encode 3D shapes into a latent space, followed by training a diffusion model on this latent space to generate 3D shapes~\cite{jun2023shap, zhang20233dshape2vecset, zhao2023michelangelo, hong20243dtopia, wu2024direct3d}. 
Differently, others aim to train 3D diffusion models  by first fitting 3D representations~\cite{zanfir2020neural, kerbl3Dgaussians} to obtain a neural representation of 3D datasets and then applying a 3D diffusion model to generate these representations~\cite{luo2021diffusion, zhou20213d, nichol2022point, wang2022rodin, jun2023shap, shue20233d, ntavelis2023autodecoding, zhang2024gaussiancube}.
In this paper, we adopt GaussianCube~\cite{zhang2024gaussiancube} and TRELLIS~\cite{xiang2024structured} as our backbone due to their strong performance in text-to-3D generation and their representativeness of 3D diffusion model capabilities.

\noindent\textbf{Inference-Time Scaling.}
Inference-time scaling refers to the strategy of utilizing additional computational resources during inference to enhance generation performance~\cite{snell2024scaling,wu2024inference,liu2025can,guo2025deepseek,zhang2025reinforced, qu2025ttom}.
Initially developed in large language models, early studies primarily focused on improving search algorithms~\cite{gandhi2024stream,su2024dualformer,wei2022chain,xie2024monte} and refining search verifiers~\cite{cobbe2021training,li2024process,lightman2023let,wang2023math}.
In the field of image generation, early exploration centered on Sequential Monte Carlo-based methods~\cite{trippe2023practical,kim2025test,yoon2025psi,skreta2025feynman,singhal2025general}.
These approaches iteratively update particles through reward signals, balancing fidelity to the pretrained model with alignment toward desired outputs.
More recently, search-based methods have demonstrated notable performance improvements for 2D diffusion models~\cite{ma2025inference,guo2025can,ramesh2025test}.
These methods enhance generation quality by explicitly searching the noise seeds~\cite{ma2025inference} or noise trajectories~\cite{ramesh2025test} guided by verifier feedback.
For example, Ma~\etal~\cite{ma2025inference} introduces a general framework that scales diffusion models at inference time by searching for better initial noise seeds using verifier feedback and search algorithms.
For a broader perspective, we refer readers to the tutorial~\cite{uehara2025inference}.
Despite these advancements in 2D diffusion models, inference-time scaling in 3D diffusion models remains unexplored.
In this work, we take a step toward inference-time scaling for text-to-3D generation.
To address the increased search dimensionality inherent in 3D tasks, we introduce a compressed search space, termed the singular search space, which enables effective and scalable inference-time exploration.

\section{Preliminaries}
\label{sec:pre}
In this section, we formally introduce the inference-time scaling problem in 3D diffusion models. 
Specifically, we begin by introducing 3D diffusion models, where we choose  GaussianCube~\cite{zhang2024gaussiancube} and TRELLIS~\cite{xiang2024structured} as the baseline for experiments.
Subsequently, we define inference-time scaling in 3D diffusion models.
Finally, we formulate the search space associated with 3D diffusion models.

\textbf{3D diffusion models.} Diffusion models are a class of generative models that iteratively transform Gaussian noise into structured data through a series of denoising steps~\cite{luo2024deem,sohl2015deep,ho2020denoising,song2020denoising,wang2024lavin}, which have been widely used in the generation of audio~\cite{kong2020diffwave,liu2023audioldm}, image~\cite{ho2022imagen,stable_diffusion}, video~\cite{ho2022video,blattmann2023stable}, \etc. 
For 3D data, recent methods~\cite{luo2021diffusion,zhou20213d,nichol2022point,wang2022rodin,jun2023shap,shue20233d,ntavelis2023autodecoding} employ 3D diffusion models to generate 3D representations. In this paper, we adopt GaussianCube~\cite{zhang2024gaussiancube} and TRELLIS~\cite{xiang2024structured} as our backbone due to their strong performance in text-to-3D generation and their representativeness of 3D diffusion model capabilities.
The generation process gradually denoises a sample starting from pure Gaussian noise $\mathbf{x}_T \in \mathcal{N}(0, \mathbf{I})$ until reaching a 3D object $\mathbf{x}_0$.

\textbf{Problem setup: Inference-time scaling.}
Inference-time scaling refers to the strategy of utilizing additional computational resources during inference to enhance generation performance~\cite{snell2024scaling,wu2024inference,liu2025can,guo2025deepseek}.
% In diffusion models, this often involves increasing the number of denoising steps or employing search algorithms to identify optimal noise.
% Recent research has demonstrated that such strategies can lead to substantial improvements in text-to-image generation~\cite{ma2025inference,guo2025can}. 
In this paper, we investigate inference-time scaling by optimizing the initial noise selection for the 3D diffusion. The algorithms used to update noise candidates and the verifiers employed to identify optimal candidates are analyzed. By carefully exploring them, we show the effectiveness of inference-time scaling laws in the 3D diffusion model~\cite{zhang2024gaussiancube, xiang2024structured}.

\textbf{Definition and formulation of search space.}
We define the search space as the initialization parameters of 3D diffusion models.
Specifically, we focus on the initial 3D Gaussian noise tensor, denoted as $\mathbf{x}_T \in \mathbb{R}^{N_\text{v} \times N_\text{v} \times N_\text{v} \times C}$, where $N_\text{v}$ is the latent resolution and $C$ is the channel number of latent feature. 
% channel number of a 3D Gaussian point~\cite{kerbl3Dgaussians}.
This Gaussian noise serves as the starting point for the reverse diffusion process.
By initializing $\mathbf{x}_T$ with normal Gaussian noise and systematically exploring variations within this space during inference, we aim to identify optimal noise that lead to higher-quality 3D assets generation.

\section{Scaling at Inference Time}
\label{sec:method}

\begin{figure*}[t]
    \centering
    \includegraphics[width=1.0\linewidth]{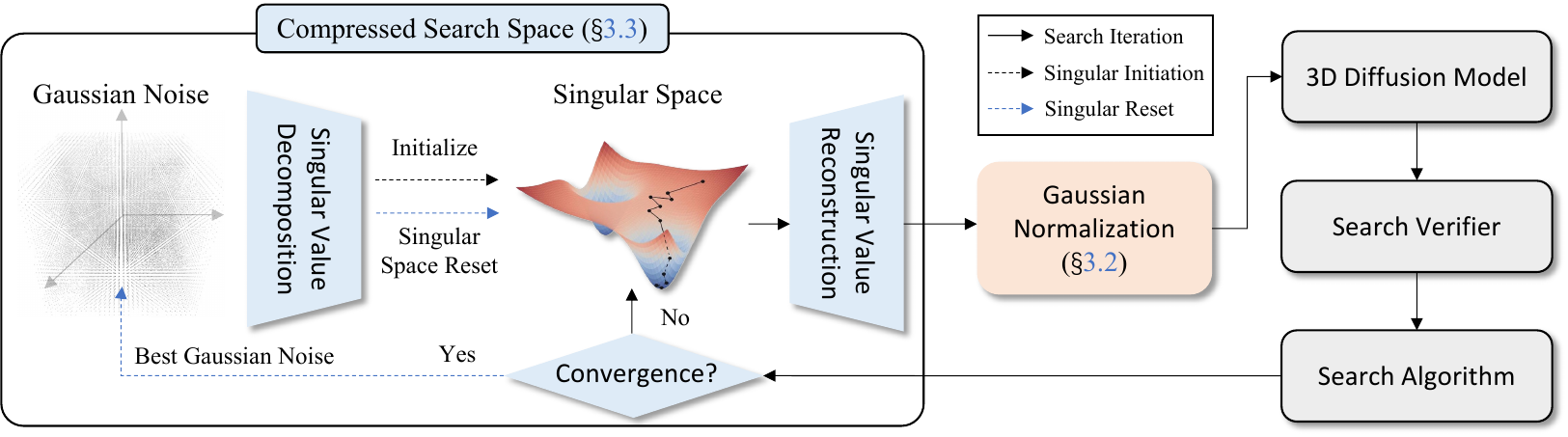}
    \vspace{-1.em}
    \caption{
    \textbf{Overview of our inference-time scaling framework for text-guided 3D diffusion models.} 
    (1) \textit{Compressed search space}: Gaussian noise undergoes SVD to initialize a lower-dimensional singular space, improving search efficiency. 
    A singular space reset mechanism updates the search space when candidate diversity decreases, preventing convergence to local minima. 
    (2) \textit{Gaussian normalization}: The optimized noise is then reconstructed and passed through Gaussian normalization to maintain a standard normal distribution, stabilizing the search process. 
    (3) \textit{Iterative search process}: The refined noise is fed into a 3D diffusion model, evaluated by a search verifier, and iteratively refined by a search algorithm, ensuring improvement through inference-time scaling.
    }
    \label{fig:overview}
    \vspace{-1.em}
\end{figure*}

Inference-time scaling can be seen as an optimization problem that finds the 3D asset with the highest evaluation score.
Below, we first introduce the search algorithm and verifier (Section~\ref{sec:method_algorithm_verifier}), then discuss the Gaussian normalization technique (Section~\ref{sec:method_normalization}) and the compressed search space (Section~\ref{sec:method_space}). The overall procedure is illustrated in Figure~\ref{fig:overview}, with a detailed algorithm flow in Appendix~\ref{app:algorithm_flow}.

\subsection{Search Verifier and Algorithm } \label{sec:method_algorithm_verifier}
This optimization process for the inference-time scaling involves two key components: the search verifier and the algorithm. 
The search algorithm iteratively updates the candidate noise based on the feedback from the search verifier, which evaluates the quality of the generated 3D asset.

\textbf{Search verifier.}
The choice of verifiers deeply influences the search results. 
We investigate two types of verifiers, including supervised and self-supervised verifiers.

\textit{Supervised verifier.}
We adopt image-aware human preference models as supervised verifiers, including ImageReward~\cite{xu2024imagereward}, HPSv2~\cite{wu2023human}, and Reward3D~\cite{ye2025dreamreward}.
These models predict human preference scores for generated images, estimating the likelihood that a given image would be favored by human evaluators~\cite{wu2023human}.
Notably, we choose HPSv2 as the default verifier without further mention.

\textit{Self-supervised verifier.}
In addition to external reward models, we explore intrinsic reward signals derived from the generative model itself.
Inspired by Feynman-Kac formulations~\cite{skreta2025feynman}, we consider the importance weight of a sampling path, denoted by a log-weight $w_t$.
Under classifier-free guidance (CFG), its incremental update is given by:
\begin{equation}
    dw_t = \frac{\sigma_t^2}{2} \cdot \beta(1 - \beta) \cdot \left\|  \nabla \log q^1_t(\mathbf{x}_t) -  \nabla\log q^2_t(\mathbf{x}_t) \right\|^2 \, dt,
    \label{eq:fk-base}
\end{equation}
where $\nabla \log q^1_t(\mathbf{x}_t)$ and $\nabla \log q^2_t(\mathbf{x}_t)$ denote the score functions of the unconditional and conditional distributions respectively, $\sigma_t$ is the noise scale at time $t$, and $\beta$ is the guidance interpolation factor.
To adapt this formulation to rectified flow models (\eg, TRELLIS~\cite{xiang2024structured}), we replace the score function with the predicted velocity field $v_t(\mathbf{x})$. 
Specifically, we make use of the equivalence $\nabla \log q_t(\mathbf{x}_t) = \frac{1}{\sigma_t^2} \left( v_t(\mathbf{x}_t) + f_t(\mathbf{x}_t) \right)$, where $f_t(\mathbf{x}_t)$ denotes the drift of the forward SDE used during training. Substituting this into the original log-weight formulation yields:
\begin{equation}
    dw_t = \frac{\beta(1 - \beta)}{2\sigma_t^2} \cdot \left\| v_t^1(\mathbf{x}_t) - v_t^2(\mathbf{x}_t) \right\|^2 \, dt,
\end{equation}
where $v^1_t(\mathbf{x}_t)$ and $v_t^2(\mathbf{x}_t)$ are the velocity fields predicted by the model under the unconditional and conditional contexts, respectively.
This self-induced score functions as an implicit reward model, providing a self-supervised signal that guides the inference-time search process.

\begin{wrapfigure}{r}{0.5\textwidth}
    \centering
    \vspace{-0.6mm}
    \includegraphics[width=0.5\textwidth]{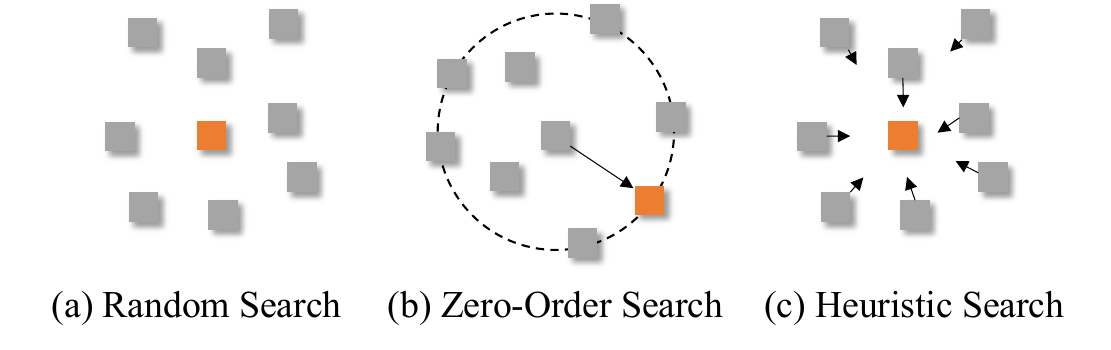}
    \vspace{-5mm}
    \caption{\textbf{Illustrations of search algorithms.}
    (a) Random search explores the search space by randomly sampling candidate points without considering prior evaluations. 
    (b) Zero-order search iteratively refines the search by selecting the best candidate from a set of perturbed samples within a defined radius (dashed circle), updating the search pivot accordingly. 
    (c) Heuristic search guides candidates toward higher-scoring regions based on heuristic-based movement strategies, such as attraction mechanisms~\cite{yang2009firefly}.}
    \label{fig:search_algorithm}
    \vspace{-3mm}
\end{wrapfigure}

\textbf{Search algorithm.}
As illustrated in Figure~\ref{fig:search_algorithm}, we investigate three search strategies, which include \textit{random search}, \textit{zero-order search}, and \textit{heuristic search} (\eg, the firefly algorithm~\cite{yang2009firefly}). Both the flows of the search strategies and pseudo codes are provided in Appendix~\ref{app:algorithm_flow}.

\textit{Random search.} As for this search strategy, we randomly sample $N$ Gaussian noises as candidates. Samples are generated from these noises. The one yielding the highest score from the verifiers is selected as the final result.

\textit{Zero-order search.} 
Zero-order search is an optimization technique that refines solutions without relying on explicit gradient information. 
It operates by initializing a random Gaussian noise as a pivot and then iteratively sampling neighboring candidates within a specified search radius. 
The candidate with the highest evaluation score, such as a preference score from a verifier, is selected to update the pivot. 
This process continues for $K$ iterations, progressively refining the search towards an optimal solution.
Formally, given a pivot $\mathbf{x}_t$ at iteration $t$, zero-order search generates $N$ perturbed candidates $\mathbf{x}_t'$ within a neighborhood radius $\lambda$.
The best-performing candidate is selected to update the pivot for the next iteration:
\begin{equation}
    \mathbf{x}_{t+1} = \arg\max_{\mathbf{x}_t'} \mathcal{F}(\mathbf{x}_t'),
\end{equation}
where $\mathcal{F}(\cdot)$ is the evaluation function, such as a verifier measuring human preference.
Here, we fix the number of candidates $N$ as a constant, and regard $K$ and $\lambda$ as scaling parameters.

\textit{Heuristic search.} Inspired by the flashing behavior of fireflies in nature, we regard each candidate noise as a firefly whose brightness is determined by the score of the verifier~\cite{yang2009firefly}. 
At each iteration, fireflies with higher brightness attract the others, causing them to move closer in noise space. This movement is governed by an attractiveness parameter and decreases with distance, balancing exploration and exploitation, which can be formulated as:
\begin{equation}
    \mathbf{x}_i = \mathbf{x}_i + \beta (\mathbf{x}_j - \mathbf{x}_i) + \alpha \bm{\epsilon},
\end{equation}
where $(\mathbf{x}_i, \mathbf{x}_j)$ are the positions of fireflies $i$ and $j$ in search space, $\alpha$ is a randomization parameter, and $\bm{\epsilon}$ is a random tensor sampled from a uniform or Gaussian distribution. Note that $\beta$ is the attractiveness factor, which follows $\beta=\beta_0 e^{-\gamma r_{ij}^2}$,
where $\beta_0$ is the initial attractiveness, $\gamma$ is the light absorption coefficient and $r_{ij} = \sqrt{\sum_{k=1}^d(\mathbf{x}_{i,k} - \mathbf{x}_{j,k})^2}$ is the Cartesian distance between two fireflies $(i, j)$.
We define $N$ fireflies at the start and run the process for $K$ iterations, treating these as the primary scaling parameters.  
Through repeated attraction and movement, the swarm collectively converges toward the firefly with the highest brightness (\textit{i.e.}, the best score).

\subsection{Gaussian Normalization} \label{sec:method_normalization}

\begin{figure*}[t]
    \centering
    \includegraphics[width=1.0\linewidth]{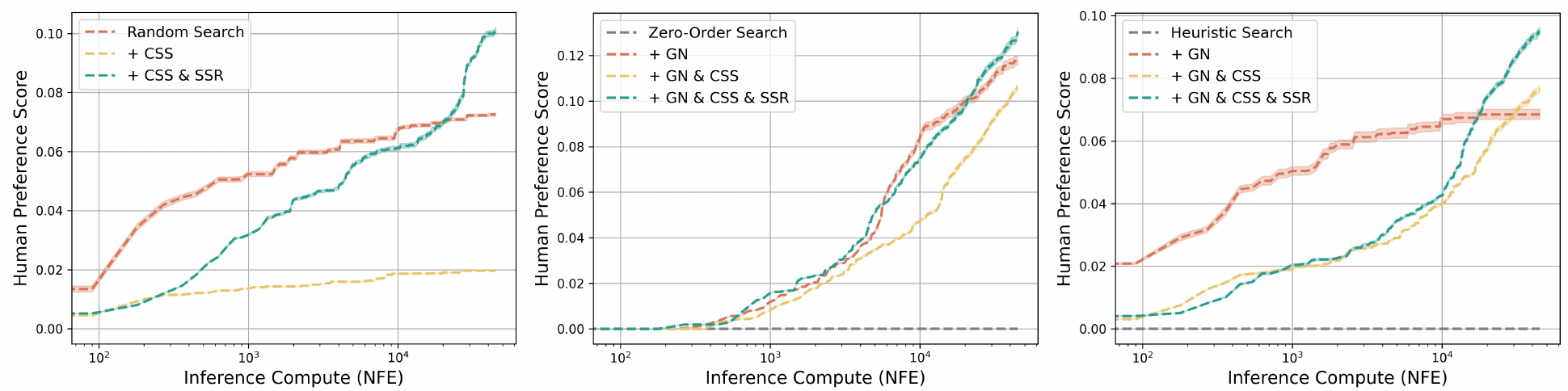}
    \vspace{-1.0em}
    \caption{
    \textbf{Ablation study on Gaussian normalization and compressed search space.}
    We evaluate the effect of Gaussian normalization and compressed search space on random search (\textit{left}), zero-order search (\textit{middle}), and heuristic search (\textit{right}).
    Across all search strategies, Gaussian normalization (termed as GN) improves stability and search effectiveness, while compressed search space (termed as CSS) with singular space reset (termed as SSR) further enhances efficiency and generation quality. 
    }
    \vspace{-1em}
    \label{fig:abl_compressed}
\end{figure*}

\noindent\textbf{Distribution shift in searching.}
In the initial experiments (refer to gray dashed lines in Figure~\ref{fig:abl_compressed}), we observe a degeneration in iterative search algorithms (\eg, zero-order search and heuristic search), where the updated candidates consistently perform worse.
We attribute this issue to a distribution shift problem. 
Specifically, the Gaussian noise cannot maintain its normal distribution during updates in the search process. 
Additionally, the generative model is sensitive to changes in the distribution of input noise. 
As a result, the cumulative shift leads to non-normal Gaussian noise, which negatively impacts the generative quality.

\textbf{Gaussian normalization layer.}
Inspired by widely used normalization techniques in deep model design~\cite{ba2016layer, wu2018group}, we introduce a Gaussian normalization layer before the 3D diffusion model.
This layer transforms the initial Gaussian noise into a normal Gaussian distribution by subtracting its mean and dividing it by its variance:
\begin{equation}
    \mathbf{x}'_T = \frac{\mathbf{x}_T - \text{Mean}[\mathbf{x}_T]}{\text{Var}[\mathbf{x}_T]}.
\end{equation}
This operation ensures that the initial Gaussian noise remains within a standard normal distribution. 
Further experiments confirm that it effectively mitigates distribution shift, enabling iterative search algorithms to function reliably (refer to red dashed lines in Figure~\ref{fig:abl_compressed}). 
Notably, it significantly enhances their performance, allowing them to surpass random search. 
This finding may also explain the key insight from~\cite{ma2025inference}, where random search outperformed iterative search algorithms—likely due to the absence of a Gaussian normalization operation.

\subsection{Compressed Search Space} \label{sec:method_space}
The search space for 3D generation is vast and high-dimensional, making it computationally expensive and time-consuming to explore thoroughly.
It causes a contradiction between the large size of the search space and the limited search budget.
To overcome this, it is essential to compress the search space to make the optimization process more effective. To this end, we first sample a Gaussian noise $\mathbf{x}_T \in \mathbb{R}^{N_\text{v} \times N_\text{v} \times N_\text{v} \times C}$ from the search space as a pivot. Afterward, we conduct the singular value decomposition~(SVD) on $\mathbf{x}_T$ to define a compressed and effective search space.

\noindent\textbf{Noise decomposition.}
We first reshape $\mathbf{x}_T$ into a batch 2D matrix with shape of $(C_\text{s}, N_\text{s}, N_\text{s})$.
Then, we perform singular value decomposition to $\mathbf{x}_T$, leading to $\mathbf{x}_T = \mathbf{U}\mathbf{\Sigma}\mathbf{V}^\top$, where $\mathbf{\Sigma}$ is a diagonal matrix containing the singular values of $\mathbf{x}_T$, and $\mathbf{U}$ and $\mathbf{V}$ involves the left and right singular vectors of $\mathbf{x}_T$, respectively.

\noindent\textbf{Search on singular values.}
Prior work~\cite{zhou2024golden} finds that the effective noise candidates share a remarkable similarity in the singular vectors in the 2D diffusion model.
Building upon this insight, we extend the observation to 3D diffusion models (see Section~\ref{sec:exp-vect_sim} for details) and explore this property to design a more effective search space.
Specifically, we fix singular vectors and only optimize singular values $\sigma_\text{init}$ (the diagonal values of $\mathbf{\Sigma} \in \mathbb{R}^{C_\text{s} \times N_\text{s} \times N_\text{s}}$), which notably reduces the search space by $N_\text{s}$ times. 
Specifically, the candidates $\left\{ \sigma_i \right\}^{N}_{i=1}$ can be sampled from $\mathcal{N}(\sigma_\text{init}, \eta\mathbf{I})$, where $\eta$ is a scaling parameter that controls the permutation.
Then, we utilize the SVD inverse transformation to reconstruct the initial Gaussian noise:
\begin{equation}
    \widetilde{\mathbf{x}}_T = \mathbf{U}\mathbf{\Sigma}_{\sigma}\mathbf{V}^\top,
\end{equation}
where $\mathbf{\Sigma}_{\sigma} = \text{diag}(\sigma)$ represents a diagonal matrix whose diagonal elements are singular values $\sigma$.

\noindent\textbf{Singular space reset.} Due to the inherent locality of iterative search algorithms (\textit{e.g.}, firefly algorithm~\cite{yang2009firefly}), candidates may converge to a local optimum, which adversely impacts search performance. 
To alleviate this issue, we introduce resetting the singular space. Specifically, at each iteration, we assess whether the search has reached a local optimum by computing the variance of the search candidates. If the variance falls below a predefined threshold $\zeta = 0.001$, we regard the best candidate as the new pivot and perform SVD to update the singular value and vectors.
Furthermore, a singular space reset can also be performed when obtaining an optimal result.
It ensures the search process remains within the singular space of the best noise and allows it to extend to random search.
This reset process redefines the search space, mitigating the local optimum and further enhancing scaling performance~(see empirical evidence in Figure~\ref{fig:abl_compressed}).
\section{Empirical Evaluations}
\label{sec:exp}

\begin{figure*}[t]
    \centering
    \includegraphics[width=1.0\linewidth]{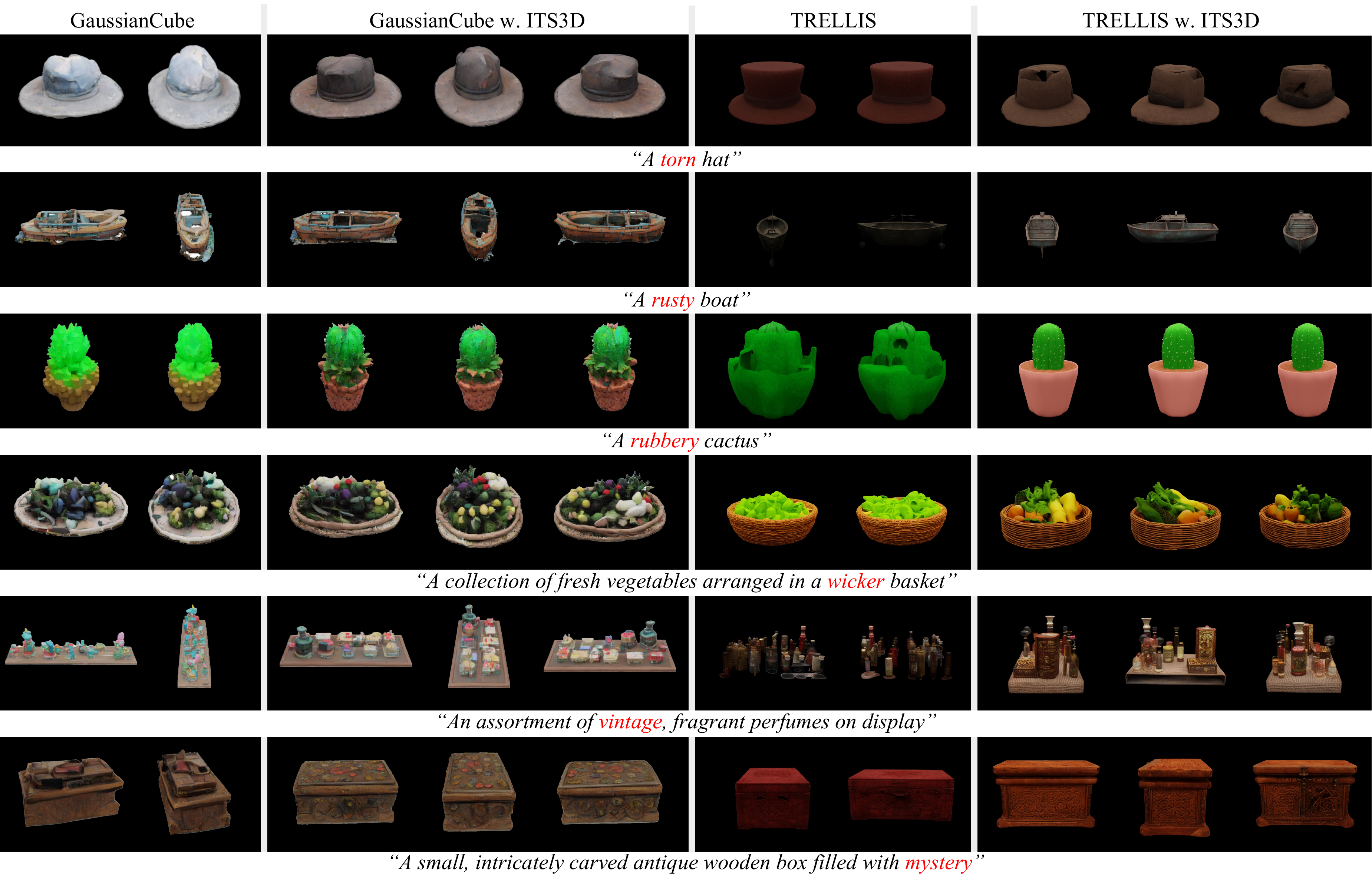}
    \vspace{-1.5em}
    \caption{
    \textbf{Qualitative comparisons on the GPTEval3D benchmark~\cite{wu2024gpt}.} 
    We compare the baselines~\cite{zhang2024gaussiancube,xiang2024structured} (\textit{left}) and the baselines with inference-time scaling (\textit{right}) across multiple prompts. 
    The baseline results often exhibit artifacts, a lack of detail, or inconsistencies in shape and texture. 
    As a comparison, applying inference-time scaling can lead to higher fidelity, improved textures, and better structural coherence.}
    \vspace{-1.5em}
    \label{fig:comp}
\end{figure*}

In this section, a series of experiments is conducted to justify our claims.
We first detail experiment setups (Section~\ref{sec:exp_setups}).
A comprehensive analysis has been conducted to examine the inference-time scaling capability in text-to-3D generation (Section~\ref{sec:exp_analysis}).
Finally, we carry out ablation studies to further elaborate on and discuss the effectiveness of our design (Section~\ref{sec:exp_ablation}).
Additional results are provided in Appendix~\ref{app:results}.

\begin{table*}[t] 
    \centering
    \small
    % \vspace{-2.0em}
     \caption{
    \textbf{Qualitative comparisons on 110 prompts generated by GPTEval3D~\protect\cite{wu2024gpt}.}
    We calculate the ImageReward score (IR)~\protect\cite{xu2024imagereward} for human preference evaluation, the CLIP score~\protect\cite{radford2021clip} for text-image alignment evaluation, and GPTEval3D~\protect\cite{wu2024gpt} for comprehensive 3D quality evaluation. The best performance in each case is shown in bold.
    }
    \resizebox{\textwidth}{!}{
    \begin{tabular}{lcccccccc} 
    \toprule
    \multirow{2}{*}{Method} & \multirow{2}{*}{IR $\uparrow$}  & \multirow{2}{*}{CLIP $\uparrow$} & \multicolumn{6}{c}{GPTEval3D $\uparrow$} \\ 
    \cmidrule{4-9}
    & & & \small{Alignment} &  \small{Plausibility} & \small{T-G Coherency} &  \small{Geo Details} &   \small{Tex Details} &  \small{Overall} \\
    \midrule
    \multicolumn{1}{l}{\textcolor{gray}{\textit{Optimization-Based Methods}}} & \multicolumn{8}{l}{} \\
    DreamFusion~\cite{poole2022dreamfusion} & -1.51 & 0.228 & 1000.0 & 1000.0 & 1000.0 & 1000.0 & 1000.0 & 1000.0 \\
    DreamGaussian~\cite{tang2023dreamgaussian} & -1.56 & 0.207 & 1100.6 & 953.6 & 1158.6 & 1126.2 & 1130.8 & 951.4 \\
    Fantasia3D~\cite{chen2023fantasia3d} & -1.40 & 0.231 & 1067.9 & 891.9 & 1006.0 & 1109.3 & 1027.5 & 933.5 \\
    Latent-NeRF~\cite{metzer2023latent} & -0.42 & 0.263 & 1222.3 & 1144.8 & 1156.7 & 1180.5 & 1160.8 & 1178.7 \\
    Magic3D~\cite{lin2023magic3d} & -1.11 & 0.243 & 1152.3 & 1000.8 & 1084.4 & 1178.1 & 1084.6 & 961.7 \\
    ProlificDreamer~\cite{wang2023prolificdreamer} & -0.50 & 0.262 & 1261.8 & 1058.7 & 1152.0 & 1246.4 & 1180.6 & 1012.5 \\
    SJC~\cite{wang2023score} & -0.82 & 0.252 & 1130.2 & 995.1 & 1033.5 & 1079.9 & 1042.5 & 993.8 \\
    SyncDreamer~\cite{liu2023syncdreamer} & -1.77 & 0.204 & 1041.2 & 968.8 & 1083.1 & 1064.2 & 1045.7 & 963.5 \\
    Wonder3D~\cite{long2024wonder3d} & -1.70 & 0.205 & 985.9 & 941.4 & 931.8 & 973.1 & 967.8 & 970.9 \\
    MVDream~\cite{shi2023mvdream} & -0.58 & 0.247 & 1270.5 & 1147.5 & 1250.6 & 1324.9 & 1255.5 & 1097.7 \\
    DreamDPO~\cite{zhou2025dreamdpo} & -0.35 & 0.259 & 1298.9 & 1171.9 & 1276.4 & 1373.2 & 1296.9 & 1203.1 \\
    \midrule
    \multicolumn{1}{l}{\textcolor{gray}{\textit{Feed-Forward-Based Methods}}} & \multicolumn{8}{l}{} \\
    Point-E~\cite{nichol2022point} & -2.24 & 0.169 & 725.2 & 689.8 & 688.6 & 715.7 & 745.5 & 618.9 \\
    Shap-E~\cite{jun2023shap} & -2.10 & 0.191 & 842.8 & 842.4 & 846.0 & 784.4 & 862.9 & 843.8 \\
    LGM~\cite{tang2024lgm} & - & - & 1091.49 & 1020 & 1108.01 & 1128.85 & 1122.08 & - \\
    \noalign{\vskip 0.25em}
    \hdashline
    \noalign{\vskip 0.25em}
    GaussianCube~\cite{zhang2024gaussiancube} & -1.84 & 0.195 & 984.72 & 938.61 & 1016.42 & 938.61 & 1056.23 & 888.96 \\
    \rowcolor{my_gray}
    \quad + ITS3D (2D Verifier~\cite{wu2023human}) & -1.39 & 0.226 & 1123.27 & 1046.17 & 1142.27 & 1144.09 & 1171.25 & 1041.89 \\
    \rowcolor{my_gray}
    \quad \% Improvement & {\color{my_green}+0.45} & {\color{my_green}+0.031} & {\color{my_green}+138.55} & {\color{my_green}+107.56} & {\color{my_green}+125.85} & {\color{my_green}+205.48} & {\color{my_green}+115.02} & {\color{my_green}+152.93} \\
    \rowcolor{my_gray}
    \quad + ITS3D (3D Verifier~\cite{ye2025dreamreward}) & -0.69 & 0.215 & 1087.45 & 1037.72 & 1121.48 & 1110.63 & 1142.40 & 1016.65 \\
    \rowcolor{my_gray}
    \quad \% Improvement & {\color{my_green}+1.15} & {\color{my_green}+0.020} & {\color{my_green}+102.73} & {\color{my_green}+99.11} & {\color{my_green}+105.06} & {\color{my_green}+172.02} & {\color{my_green}+86.17} & {\color{my_green}+127.69} \\
    TRELLIS~\cite{xiang2024structured} & -1.81 & 0.224 & 1085.17 & 1083.49 & 1124.51 & 1109.71 & 1136.59 & 1039.65 \\
    \rowcolor{my_gray}
    \quad + ITS3D (2D Verifier~\cite{wu2023human}) & -0.814 & \textbf{0.251} & 1206.99 & 1228.62 & 1265.76 & 1236.56 & 1266.32 & 1180.45 \\
    \rowcolor{my_gray}
    \quad \% Improvement & {\color{my_green}+0.996} & {\color{my_green}+0.027} & {\color{my_green}+121.82} & {\color{my_green}+145.13} & {\color{my_green}+141.25} & {\color{my_green}+126.85} & {\color{my_green}+129.73} & {\color{my_green}+140.80} \\
    \rowcolor{my_gray}
    \quad + ITS3D (3D Verifier~\cite{ye2025dreamreward}) & \textbf{-0.14} & \textbf{0.251} & \textbf{1248.05} & \textbf{1243.64} & \textbf{1290.61} & \textbf{1243.59} & \textbf{1279.08} & \textbf{1223.77} \\
    \rowcolor{my_gray}
    \quad \% Improvement & {\color{my_green}+1.67} & {\color{my_green}+0.027} & {\color{my_green}+162.88} & {\color{my_green}+160.15} & {\color{my_green}+166.10} & {\color{my_green}+133.88} & {\color{my_green}+142.49} & {\color{my_green}+184.12} \\
    \bottomrule
    \end{tabular}}
    \label{tab:qua_comp}
    \vspace{-1.0em}
\end{table*}

\subsection{Experimental Setups} \label{sec:exp_setups}
\textbf{Datasets and measurements.}
We mainly focus on text-to-3D generation.
To this end, we choose GPTEval3D~\cite{wu2024gpt} as the default dataset.
It contains 110 prompts that cover a range of creativity and complexity use cases.
Based on this, we adopt three commonly used evaluation strategies.
(1)~\textit{CLIPScore}~\cite{radford2021clip} assesses the image-text alignment by averaging clip scores from multi-view rendering from a 3D asset and the given text prompt. 
(2)~\textit{ImageReward}~\cite{xu2024imagereward} measures the human preference alignment between multi-view renderings and the given text prompt.
(3)~\textit{GPTEval3D}~\cite{wu2024consistent3d} performs pairwise comparisons with baselines, generating Elo ratings that align with human judgments on text alignment, 3D plausibility, and texture-geometry coherence, \textit{etc}. 
Details of the measurements can be found in Appendix~\ref{app:metrics}.

\textbf{Baselines.}
For a comprehensive comparison, we benchmark our method against 15 baselines categorized into optimization-based~\cite{poole2022dreamfusion,tang2023dreamgaussian,chen2023fantasia3d,metzer2022latent,lin2023magic3d,shi2023mvdream,wang2023prolificdreamer,wang2023score,liu2023syncdreamer,long2024wonder3d,zhou2025dreamdpo} and feed-forward methods~\cite{nichol2022point,jun2023shap,tang2024lgm,zhang2024gaussiancube}.
Note that we use GaussianCube~\cite{zhang2024gaussiancube} and TRELLIS~\cite{xiang2024structured} as our backbone.
Details of the base models can be found in Appendix~\ref{app:base_models}.

\textbf{Implementation.} 
We conduct experiments using PyTorch~\cite{paszke2019pytorch}.
For benchmark comparisons, we use HPSv2~\cite{wu2023human}, heuristic search with Gaussian normalization, and compressed search space as the default search verifier, search algorithm, and search space, respectively.
In zero-order search, we set the default parameters as $N=5$ and $\lambda=2.0$. 
For heuristic search, we configure $N=10$ with default values $\beta=1.0$, $\gamma=0.00001$, and $\alpha=0.97$.
% For each sample, we search with 1,000 iterations, which takes around one hour on a single NVIDIA RTX A6000 GPU.
We perform 1,000 search iterations per sample, which takes approximately 30 minutes on a single NVIDIA H100-80G GPU.
On average, the search process takes 22.7 minutes for GaussianCube and 35.1 minutes for TRELLIS-Text-Large, with peak GPU memory usage of about 20 GB and 46 GB, respectively.
Given the varying denoising costs across different diffusion models, we standardize the reporting by treating each denoising step as one inference compute (NFE), where NFE stands for Number of Function Evaluations.

\begin{figure*}[t]
    \centering
    \vspace{-2.0em}
    \includegraphics[width=1.0\linewidth]{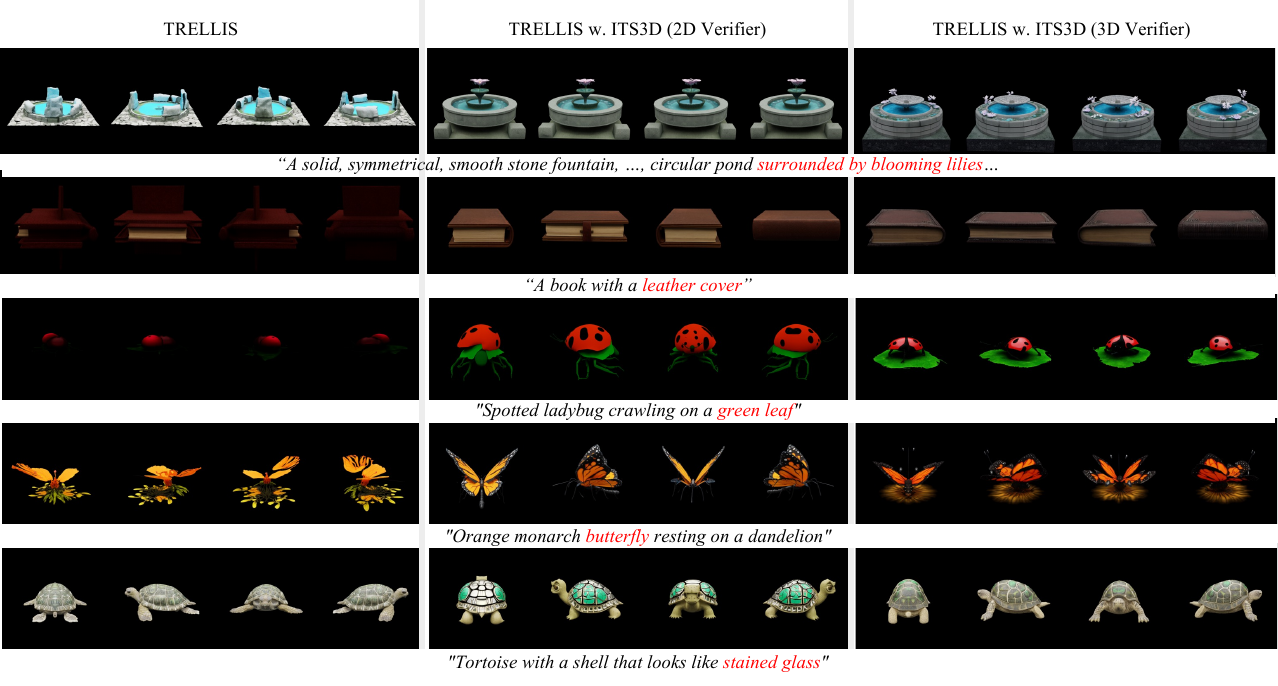}
    \vspace{-2.0em}
    \caption{
    \textbf{Ablation studies on verifiers.}
    Comparison between 3D-aware verifier Reward3D~\cite{ye2025dreamreward} and 2D-aware verifier HPSv2~\cite{wu2023human} under TRELLIS-\texttt{Large}~\cite{xiang2024structured}.
    }
    \vspace{-1.0em}
    \label{fig:abl_verifier}
\end{figure*}

\subsection{Analysis Results: Text-to-3D Generation} \label{sec:exp_analysis}

\subsubsection{Qualitative Comparisons}
To evaluate the effectiveness of inference-time scaling, we conduct qualitative comparisons between the baseline method~\cite{zhang2024gaussiancube} and the baseline with Inference-Time Scaling, as shown in Figure~\ref{fig:comp}. 
We observe that the baseline method often struggles with structural coherence, texture fidelity, and semantic alignment with the text prompts. 
For example, in the case of ``\textit{A torn hat}'' and ``\textit{A rusty boat}'', the baseline results appear blurry and lack detailed texture, whereas inference-time scaling produces assets with better prompt alignment and well-defined material properties. 
For more complex prompts, such as ``\textit{A collection of fresh vegetables arranged in a wicker basket}", the baseline method often leads to blurry or unstructured arrangements. 
In contrast, inference-time scaling improves the definition of individual vegetables, enhances color vibrancy, and maintains a more realistic composition.
These qualitative improvements demonstrate that inference-time Scaling effectively refines 3D assets, enhancing their realism, text alignment, and overall visual quality without additional training. 
By leveraging structured search and preference-based verifiers, inference-time scaling reduces artifacts, improves structural fidelity, and generates more visually appealing 3D assets.

\subsubsection{Quantitative Comparisons}
To evaluate the effectiveness of our method, quantitative comparisons on 110 prompts from GPTEval3D are provided. Table~\ref{tab:qua_comp} presents the results across multiple evaluation metrics, including ImageReward (IR) for human preference, CLIP for text-image alignment, and GPTEval3D scores for comprehensive 3D quality assessment. Specifically, our method significantly improves upon GaussianCube, achieving a higher ImageReward score (-1.39~\vs~-1.84), which indicates better human preference alignment. 
Additionally, it achieves a higher CLIPScore (0.226~\vs~0.195), suggesting improved text-image alignment. Besides, 
in GPTEval3D metrics, our method outperforms GaussianCube across all aspects, with notable gains in text-asset alignment (+138.55), 3D plausibility (+107.56), text-geometry alignment (+125.85), geometry details (+205.48), texture details (+115.02), and overall performance (+152.93). 
These results confirm that inference-time scaling can enhance generation quality, enabling feed-forward models to narrow the performance gap with optimization-based methods.
Furthermore, they emphasize the potential of using inference-time scaling as a plug-and-play tool for improving text-to-3D generation.

\subsection{Ablation and Analysis} \label{sec:exp_ablation}
We comprehensively analyze the ITS3D framework through ablation and mechanism studies. 
We first examine the impact of key components, including Gaussian normalization, compressed search space, and singular space reset. 
We then evaluate the sensitivity of the method to hyperparameters, verifier choices, and model scales to assess its robustness and generalization. 
Finally, we provide insights into the underlying mechanisms that explain why ITS3D effectively enhances inference-time optimization.

\subsubsection{Ablation Studies on Core Components}
We first investigate the effectiveness of the key components in the ITS3D framework, including Gaussian normalization, compressed search space, and singular space reset, to understand their individual and combined contributions to search stability and performance.

\noindent\textbf{Benefit of Gaussian normalization.}
We further validate the critical role of Gaussian normalization in the search process. 
Experiments in Figure~\ref{fig:abl_compressed} show that Gaussian normalization is crucial for both zero-order search and heuristic search. 
We attribute this to the fact that generative models are sensitive to distribution drift. 
Without proper normalization, the search process may shift to a non-normal Gaussian distribution, leading to poor generation results, thus highlighting the importance of Gaussian normalization.

\textbf{Benefit of compressed search space.} 
As shown in Figure~\ref{fig:abl_compressed}, the full performance of compressed search space (green dashed line) exhibits a distinct trend across different search strategies.
As iterations increase, we observe that it gradually surpasses the baseline and achieves significantly larger improvements in later stages, especially in random search and heuristic search. 
We attribute this delayed yet stronger long-term performance to that the compressed search space effectively removes redundant and uninformative dimensions in the early stage, guiding the search towards more meaningful subspaces. 
Initially, this constraint may reduce search diversity, slowing down exploration, but over time, it filters out noise and enables more efficient refinement of candidates.  As a result, it leads to accelerated improvements in later iterations. 

\textbf{Benefit of singular space reset.} 
To assess the impact of singular space reset, we conduct an ablation study without this module, represented by the yellow dashed line in Figure~\ref{fig:abl_compressed}.
Without a singular space reset, we observe a significant performance degradation (green dashed line \vs yellow dashed line), particularly for random search. This highlights the crucial role of singular space reset in maintaining an effective compressed search space.
Moreover, this finding suggests a positive correlation between search performance and the quality of the singular space, which demonstrates their mutually reinforcing relationship.
When the singular space quality is poor, the search results degenerate, resulting in a slow initial improvement. 
By dynamically updating the singular space through singular space reset, we enhance its quality, ultimately enabling the compressed search space to achieve exponential performance gains in the later stages.

\begin{figure*}[t]
    \centering
    \includegraphics[width=1.0\linewidth]{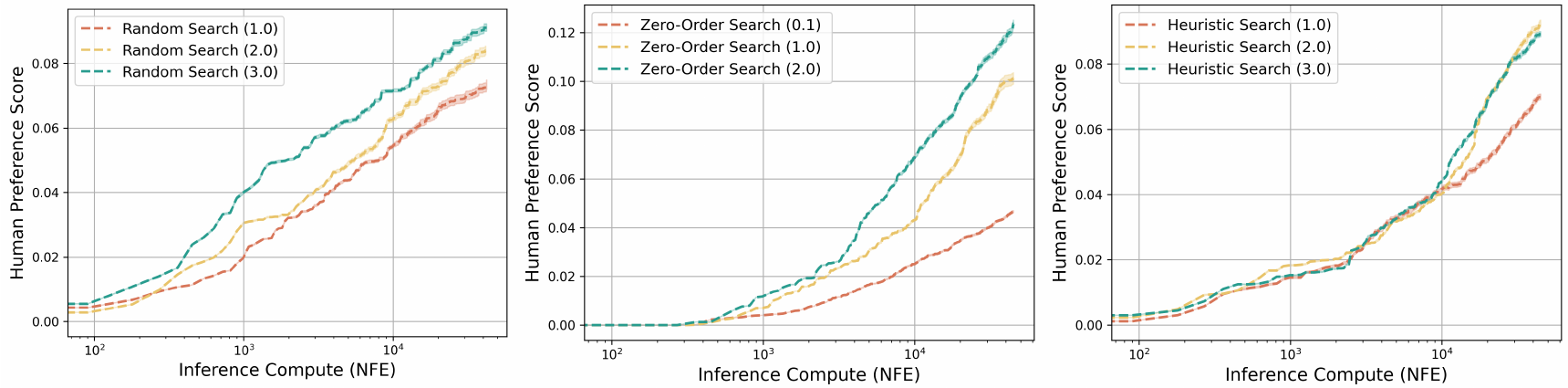}
    \vspace{-1.0em}
    \caption{
    \textbf{Ablation study on parameter variations in zero-order search and heuristic search.}
    We examine the effect of varied parameter settings on random search (\textit{left}), zero-order search (\textit{middle}), and heuristic search (\textit{right}).
    We examine the effect of permutation strength $\eta$ in compressed search space in random search and heuristic search (denoted as ``Random Search ($\eta$)'' and ``Heuristic Search ($\eta$)'') and neighborhood radius $\lambda$ in zero-order search (denoted as ``Zero-Order Search ($\lambda$)''). 
    }
    \vspace{-1em}
    \label{fig:abl_parameter}
\end{figure*}

\subsubsection{Robustness and Generalization Analysis}
We systematically analyze the sensitivity of ITS3D to different external factors, including hyperparameters, verifiers, and model scales, to assess its robustness and generalization across diverse settings.

\textbf{Ablation study on parameter variations.} We examine the effect of varied parameter settings on random search, zero-order search, and heuristic search in Figure~\ref{fig:abl_parameter}.
Specifically, we investigate the effect of neighborhood radius $\lambda$ in zero-order search and permutation strength $\eta$ in compressed search space in random search and heuristic search. 
In zero-order search, increasing the scaling parameter leads to significant improvements in search performance, with higher values yielding better results.
It indicates that a larger search radius helps discover high-quality candidates more efficiently in zero-order search.
Meanwhile, we observe that a larger permutation strength leads to better results in both random search and heuristic search (\eg, setting $\eta = 3.0$ outperforms $\eta = 1.0$).
We attribute this to the fact that strong perturbations may expand the search space, helping the search algorithm to discover better solutions.

\begin{figure*}[t]
    \centering
    \begin{minipage}[t]{0.60\textwidth}
        \centering
        \includegraphics[width=\linewidth]{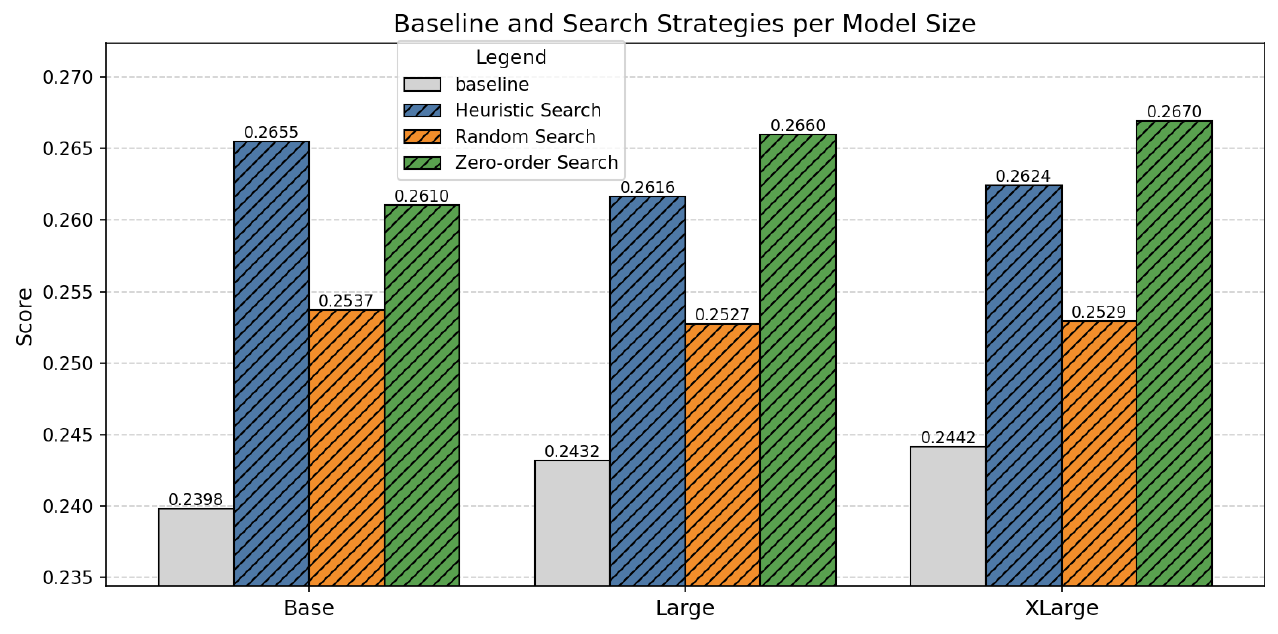}
        \caption{
        \textbf{Ablation studies on model sizes.}
        Comparison of search effectiveness across different base model sizes on TRELLIS~\cite{xiang2024structured} (\texttt{Base}, \texttt{Large}, and \texttt{XLarge}) using three search strategies (\textit{right}).
        }
        \label{fig:abl_models}
    \end{minipage}
    \hfill
    \begin{minipage}[t]{0.37\textwidth}
        \centering
        \includegraphics[width=\linewidth]{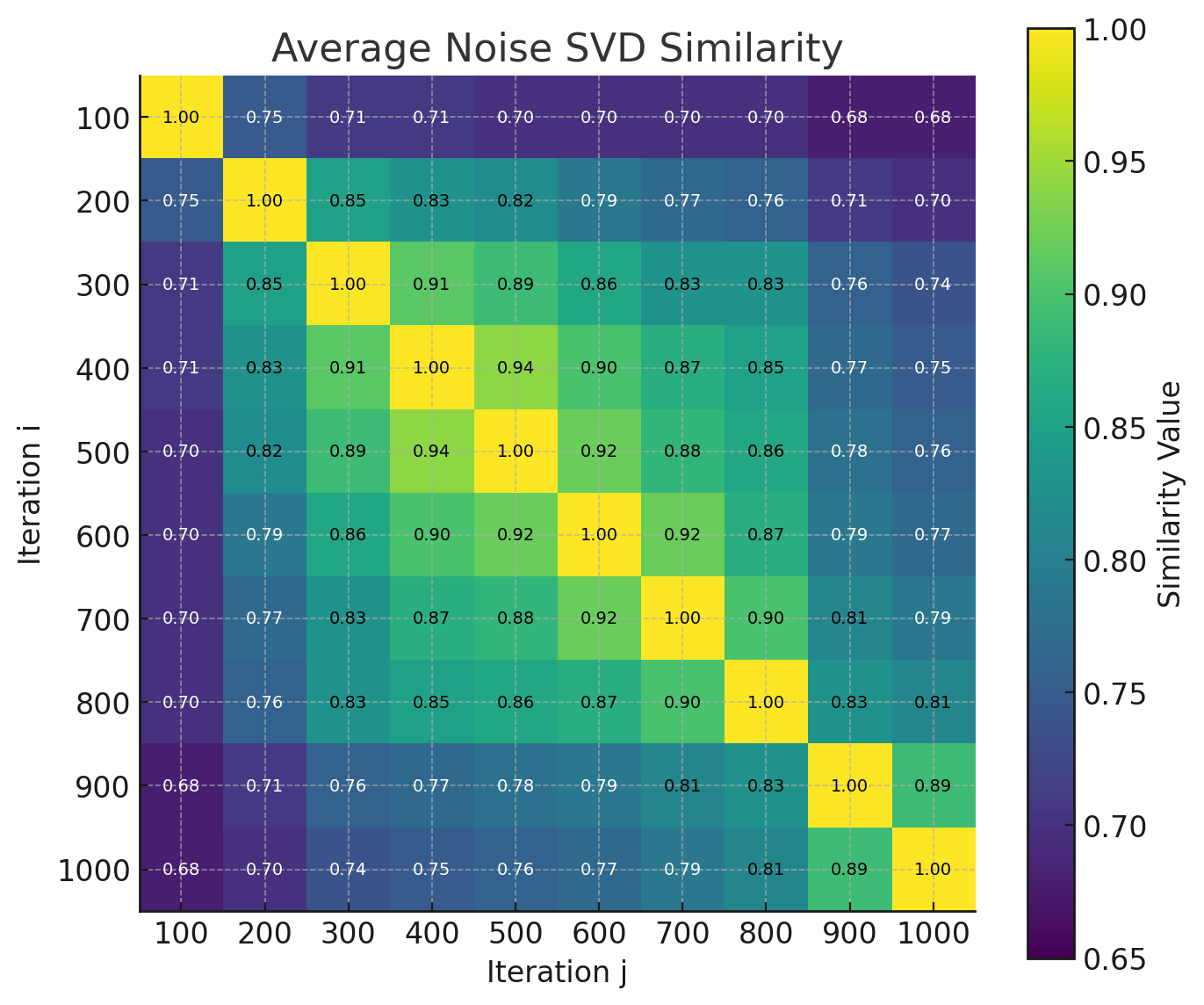}
        \caption{
        \textbf{Analysis of singular vector similarity across inference iterations.}
        Nearby iterations show high similarity.
        % (\eg, 0.94 between 400 and 500 iterations).
        }
        \label{fig:svd_anlysis}
    \end{minipage}
    \vspace{-1.0em}
\end{figure*}

\noindent\textbf{Evaluation on different base models.}
We evaluate the robustness of our method across base models of varying sizes.
Specifically, we test three search algorithms on TRELLIS~\cite{xiang2024structured} models in three configurations: \texttt{Base}, \texttt{Large}, and \texttt{XLarge}.
As shown in Figure~\ref{fig:abl_models} (\textit{right}), all three search algorithms yield performance improvements over the baseline. 
Notably, zero-order search shows strong capability in exploring model potential, achieving the best results on the \texttt{Large} and \texttt{XLarge} models, with scores of 0.2660 and 0.2670, respectively.
Meanwhile, heuristic search performs well on the \texttt{Base} model, achieving a result comparable to the best score of the \texttt{Large} model (0.2655 \vs 0.2660).
This suggests that inference-time scaling can be an effective strategy for deploying smaller models.

\begin{wraptable}{r}{0.6\textwidth}
\vspace{-1em} 
\centering
\caption{
\textbf{Sensitivity analysis to initialization quality.} 
Pearson correlation coefficients (top) and $p$-values (bottom).
Strong positive correlations are observed between initial and final scores, while correlations with improvements are generally insignificant, indicating robustness to initialization.
}
\small
\begin{tabular}{lcccc}
\toprule
 & Reward3D & Imagereward & HPSv2 & CLIP \\
\midrule
Init.~vs~Final & 
\begin{tabular}[c]{@{}c@{}}0.890\\ ($p{=}0.000$)\end{tabular} & 
\begin{tabular}[c]{@{}c@{}}0.930\\ ($p{=}0.000$)\end{tabular} & 
\begin{tabular}[c]{@{}c@{}}0.871\\ ($p{=}0.000$)\end{tabular} & 
\begin{tabular}[c]{@{}c@{}}0.686\\ ($p{=}0.000$)\end{tabular} \\
Init.~vs~Impr. & 
\begin{tabular}[c]{@{}c@{}}$-0.245$\\ ($p{=}0.192$)\end{tabular} & 
\begin{tabular}[c]{@{}c@{}}$-0.293$\\ ($p{=}0.117$)\end{tabular} & 
\begin{tabular}[c]{@{}c@{}}$-0.518$\\ ($p{=}0.003$)\end{tabular} & 
\begin{tabular}[c]{@{}c@{}}$-0.203$\\ ($p{=}0.282$)\end{tabular} \\
\bottomrule
\end{tabular}
\vspace{-1em}
\label{tab:correlation-rewards}
\end{wraptable}
\noindent\textbf{Sensitivity analysis to the initial noise distribution.}
We conducted a correlation analysis to evaluate the sensitivity of ITS3D to initialization quality across four evaluation metrics.
Specifically, we sampled 10 prompts from GPTEval3D, kept all search parameters fixed, and varied only the five random seeds used for initialization.
The results shown in Table~\ref{tab:correlation-rewards} indicate statistically positive correlations between the initial and final scores for all metrics ($r>0.68$ and $p<0.001$), indicating that better initializations consistently lead to higher final performance.
In contrast, for most metrics (\eg, Reward3D, ImageReward, CLIPScore), the correlation between initial score and relative improvement score is not statistically significant.
It suggests that ITS3D may achieve consistent relative gains regardless of initialization quality, highlighting its robustness.

\noindent\textbf{Evaluation on different supervised verifiers.} 
We conduct a quantitative analysis of the 2D-aware verifier HPSv2~\cite{wu2023human} and the 3D-aware verifier Reward3D~\cite{ye2025dreamreward}, applied to different baseline models (\eg, GaussianCube~\cite{zhang2024gaussiancube} and TRELLIS-\texttt{Large}~\cite{xiang2024structured}) on 110 prompts from the GPTEval3D benchmark~\cite{wu2024gpt}. Table~\ref{tab:qua_comp} shows that both the 2D and 3D verifiers are effective when used with inference-time scaling across different baselines. 
Among them, TRELLIS with the 3D verifier achieves the best scores on three evaluation metrics. 
While both verifiers improve performance, the 3D verifier generally outperforms the 2D version (significantly improved IR score and comparable GPTEval3D scores on GaussianCube). 
Furthermore, we qualitatively compare the effectiveness of 2D and 3D verifiers using TRELLIS-\texttt{Large}. 
Results in Figure~\ref{fig:abl_verifier} indicate that both verifiers enhance the rendering quality by improving texture and geometric details.  
They also help mitigate the Janus problem, as seen in the example ``\textit{A book with a leather cover}''.
However, in most cases, the 3D verifier produces results that better align with human preferences, particularly in prompts with highlighted key prompts (\eg, those marked in red). 
We attribute this to the ability of the 3D verifier to better capture spatial consistency and geometric plausibility, which are critical for evaluating 3D content.

\begin{wraptable}{r}{0.6\textwidth}
\vspace{-1em}
\centering
\caption{
\textbf{Evaluation on self-supervised verifier.} 
Values denote relative improvements.
The self-supervised verifier improves both CLIPScore and HPSv2, with further gains from the compressed search space.
}
\small
\begin{tabular}{@{}p{5cm}<{\centering}p{2.0cm}<{\centering}p{2.0cm}<{\centering}@{}}
\toprule
Method & CLIP & HPSv2 \\
\midrule
Zero-Order (w/o CSS) & 0.0066 & 0.0163 \\
\textbf{Zero-Order (w/ CSS)} & \textbf{0.0415} & \textbf{0.0195} \\
\bottomrule
\end{tabular}
\vspace{-1em}
\label{tab:fks-results}
\end{wraptable}
\noindent\textbf{Evaluation on self-supervised verifier.}
We evaluate the self-supervised verifier on 10 prompts sampled from GPTEval3D, using the Zero-Order search strategy.
As shown in Table~\ref{tab:fks-results}, the self-supervised verifier improves both CLIPScore and HPSv2, suggesting that the self-induced importance score could reflect the quality of generated 3D samples.
Furthermore, conducting the search within the compressed search space yields additional performance gains, demonstrating the effectiveness and scalability of the ITS3D framework.

\subsubsection{Mechanism Analysis} \label{sec:exp-vect_sim}
We further conduct an in-depth analysis of the underlying mechanisms that drive the effectiveness of ITS3D, focusing on the role of singular vector similarity and SVD-based search space compression.

\noindent\textbf{Analysis of singular vectors similarity in 3D Diffusion Model.}
We evaluate the local and global similarities of singular vectors in the 3D diffusion model (TRELLIS-Text-Large) under two settings.
1) Local Similarity.
To simulate the inference-time scaling, we construct pairs of source and target noise. Specifically, for each randomly sampled source noise, the corresponding target noise is generated by adding a small perturbation drawn from a scaled Gaussian distribution ($\lambda \mathcal{N}(0, \mathbf{I})$) to its singular values.
We ablate the effect of the perturbation strength by varying $\lambda \in \{ 0.1, 1.0\}$, and construct 100 noise pairs for each $\lambda$. 
Similar to \cite{zhou2024golden}, we compute the absolute cosine similarity between the singular vectors of the source and target noise the measure the singular vector similarity.
The average cosine similarities between singular vectors are 0.92 for $\lambda=0.1$, 0.68 for $\lambda=1.0$, and 0.66 for $\lambda=2.0$.
The results show that for small perturbations (\eg, $\lambda = 0.1$), the singular vectors remain highly consistent (cosine similarity $> 0.92$ across all dimensions). 
2) Global Similarity.
We further assess the global similarity of singular vectors across different stages of inference-time scaling.
Specifically, we randomly sample 10 prompts from the GPTEval3D benchmark and perform inference-time scaling with 5 different random seeds per prompt. During optimization, we record the best-performing noise every 100 iterations, resulting in 10 checkpoints per run and a total of 50 noise sequences.
For each sequence, we report the pairwise cosine similarity scores (average across all dimensions) between the singular vectors of the saved noises at different iterations.
As shown in Figure~\ref{fig:svd_anlysis}, the results indicate the singular vectors show strong similarities between nearby iterations (\eg, 0.94 between 400 and 500 iterations).
As optimization progresses, this similarity gradually decreases, which can be attributed to the singular space reset.

\begin{wraptable}{r}{0.6\textwidth}
\vspace{-1em}
\centering
\caption{
\textbf{Analysis of the compressed search space.} 
Performance comparison between compressed and original search spaces across different distances.
The SVD-based compressed search space yields consistently better results than the original across all search radii $\epsilon$.
}
\small
\begin{tabular}{lcccc}
\toprule
Search Space & $\epsilon{=}0.01$ & $\epsilon{=}1$ & $\epsilon{=}2$ & $\epsilon{=}3$ \\
\midrule
Vanilla Search Space & 0.2713 & 0.2708 & 0.2696 & 0.2685 \\
\textbf{Compressed Search Space} & \textbf{0.2720} & \textbf{0.2720} & \textbf{0.2719} & \textbf{0.2712} \\
\bottomrule
\end{tabular}
% \vspace{-1em}
\label{tab:search_space_comparison}
\end{wraptable}
\noindent\textbf{Analysis of the compressed search space.}
The SVD-based search space compression is motivated by the observation that effective noise candidates in 3D diffusion models tend to share similar singular vectors.
It suggests that the singular value space is semantically meaningful and can serve as a basis for search.
To validate this hypothesis, we conducted a toy experiment to assess the quality of the search space.
Specifically, we randomly initialized 10 Gaussian noises and sampled 10 prompts from GPTEval3D. 
For each noise, we simulated zero-order search by generating 10 candidates within a radius $\epsilon$, and evaluated the resulting generations from TRELLIS-Text-Large using HPSv2.
The results in Table~\ref{tab:search_space_comparison} show that the compressed search space consistently outperforms the original across various search radii $\epsilon$, suggesting it offers a more efficient and effective search space.
We attribute this advantage to SVD-based compression, which filters out irrelevant components and sharpens the search direction.
%\vspace{-2mm}
\section{Conclusion}
\label{sec:clu}
This work introduces an inference-time scaling framework for text-to-3D generation using diffusion models. 
By optimizing noise selection through structured search, Gaussian normalization, and compressed search space, our method improves the quality of generation results without additional training.
Our key contributions lie in the integration of efficient search strategies, the stabilization of iterative search through Gaussian normalization, and the compression of the search space using SVD, which collectively enhance both the effectiveness and efficiency of inference-time scaling in text-to-3D generation.
Extensive experiments demonstrate the superiority of our method over baselines across multiple metrics. 

\section*{Acknowledgements}
This research/project is supported by the National Research
Foundation, Singapore under its National Large Language Models Funding Initiative (AISG Award No: AISG-NMLP2024-002). Any opinions, findings and conclusions or recommendations expressed in this material are those of the author(s) and do not reflect the views of National Research Foundation, Singapore. Xiaobo Xia is also supported by MoE Key Laboratory of Brain-inspired Intelligent Perception and Cognition, University of Science and Technology of China (Grant No. 2421002). 

%\noindent\textbf{Limitations and future work.} In this work, we primarily focus on constructing an inference-time scaling framework for text-to-3D generation, enhancing search algorithms, and compressing the search space for improved efficiency and effectiveness. Currently, we rely on 2D-aware verifiers and extend them to 3D by averaging multi-view scores, which may not be sufficient for a comprehensive assessment of 3D quality. Therefore, the investigation of 3D-aware verifiers is important, \textit{e.g.}, UniGS~\cite{liunigs} and CLIP-GS~\cite{jiao2024clipgs}. Additionally, it is essential to integrate a semantic-aware verifier for fine-grained adjustments to 3D generation based on user inputs.  For instance, users may want specific modifications such as ``\textit{I hope the ears of the tiger are longer.}''  Developing verifiers capable of interpreting such instructions and effectively guiding 3D generation will prompt more controllable and adaptable generation results. 

% \newpage
\bibliography{main}
\bibliographystyle{unsrt}

\appendix
\clearpage
\onecolumn
\appendix

\section{Appendix}
The Appendix is organized as follows:
\begin{itemize}[leftmargin=2.5em]
    \item Appendix~\ref{app:results} presents additional qualitative results, demonstrating that both 2D and 3D verifiers improve generation quality, with the 3D verifier achieving stronger alignment with human preferences.
    
    \item Appendix~\ref{app:implemtation} details the implementation of our baseline methods, GaussianCube and TRELLIS. It explains how we incorporate them into our framework and defines the corresponding search spaces and compression strategies used during inference-time optimization.

    \item Appendix~\ref{app:add_settings} outlines additional experimental settings, including descriptions of evaluation metrics such as CLIPScore, ImageReward, and GPTEval3D. These details provide transparency and reproducibility for our experimental protocols.

    \item Appendix~\ref{sec:impact} discusses the broader societal impacts of our work, including both positive implications and potential risks. We emphasize the importance of accessibility, efficiency, and ethical considerations in the deployment of generative 3D models.

    \item Appendix~\ref{sec:reproduce} provides reproducibility information, including code availability and implementation details. We describe our commitment to open-sourcing the code and facilitating future research built on our framework.

    \item Appendix~\ref{sec:limitation} outlines current limitations of our approach, particularly regarding the need for more fine-grained and semantics-aware verifiers for controllable generation. We highlight this as a promising future direction for improving interaction and alignment with user intent.
\end{itemize}

\subsection{Additional Experiments}
\label{app:results}

\subsubsection{Additional Qualitative Results}\hfill

\begin{figure*}
    % \centering
    % \hspace*{-0.8cm}
    \includegraphics[width=1.0\textwidth]{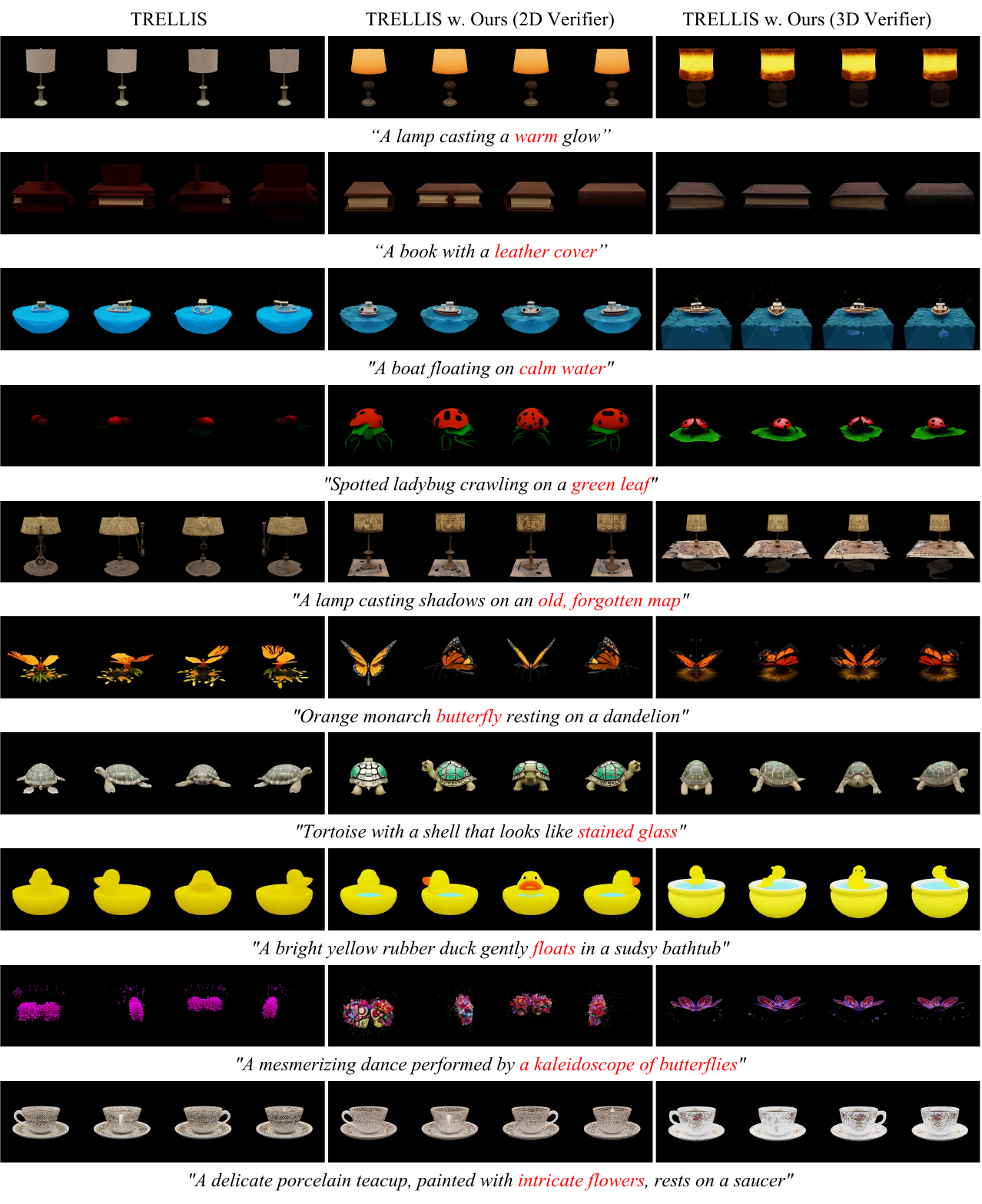}
    \captionsetup{width=1.0\textwidth}
    \caption{
    More qualitative comparisons on the GPTEval3D benchmark~\cite{wu2024gpt}, which shows results from the baseline~\cite{xiang2024structured}, and our inference-time scaling with the 2D verifier~\cite{wu2023human} and 3D verifier~\cite{ye2025dreamreward} across multiple prompts.
    }
    \label{fig:supp_vis1}
\end{figure*}

\begin{figure*}
    \centering
     % \hspace*{-0.8cm}
    \includegraphics[width=1.0\textwidth]{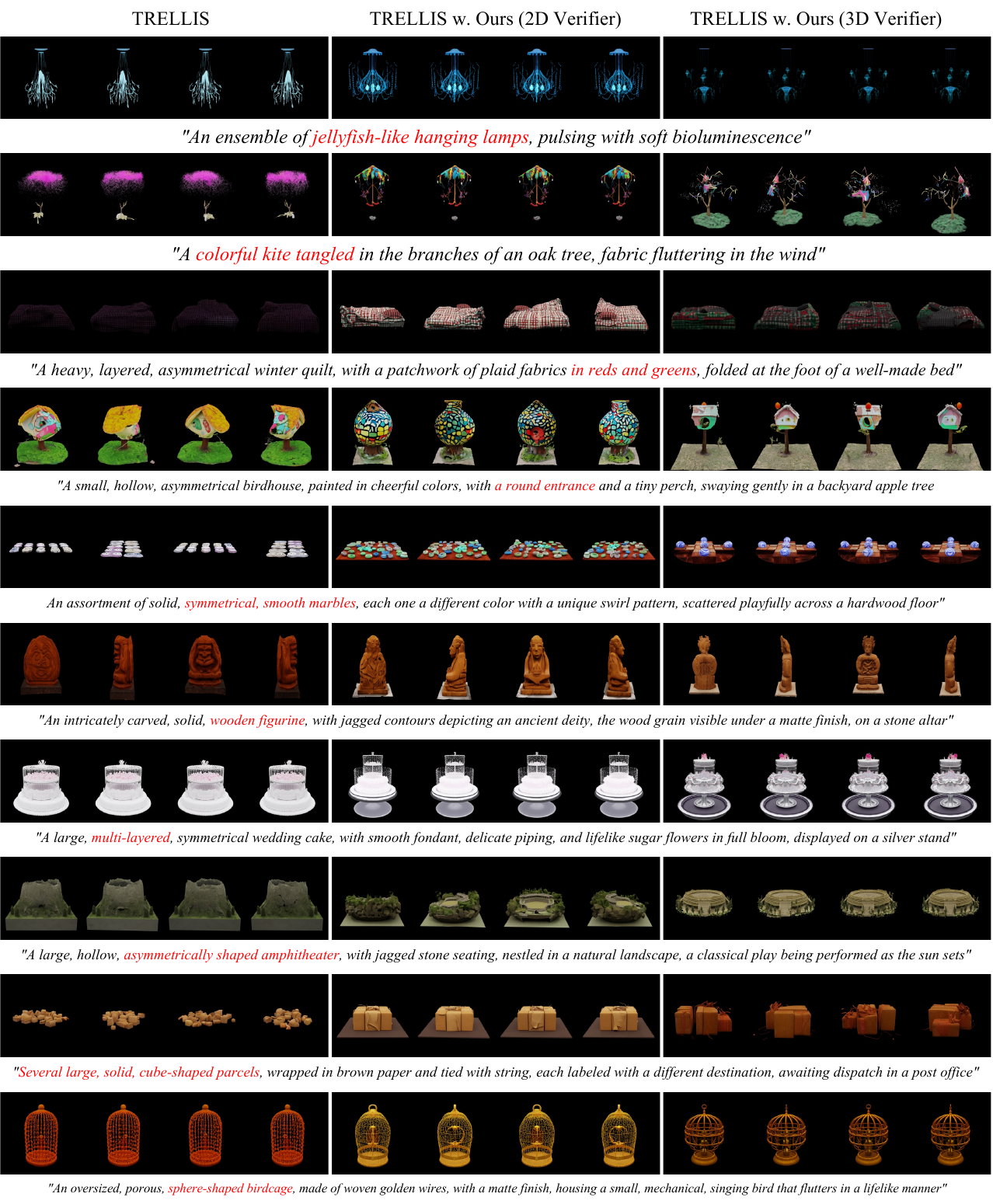}
    \captionsetup{width=1.0\textwidth}
    \caption{
    More qualitative comparisons on the GPTEval3D benchmark~\cite{wu2024gpt}, which shows results from the baseline~\cite{xiang2024structured}, and our inference-time scaling with the 2D verifier~\cite{wu2023human} and 3D verifier~\cite{ye2025dreamreward} across multiple prompts.
    }
    \label{fig:supp_vis2}
\end{figure*}

\noindent\textbf{More qualitative comparisons.}
Figure~\ref{fig:supp_vis1} and Figure~\ref{fig:supp_vis2} present qualitative comparisons between the TRELLIS baseline and our inference-time scaling method on a diverse set of prompts from the GPTEval3D benchmark~\cite{wu2024gpt}. 
The results demonstrate consistent improvements in both visual fidelity and alignment with the input text prompts. 
In examples such as ``Spotted ladybug crawling on a green leaf'' and ``A mesmerizing dance performed by a kaleidoscope of butterflies'' the vanilla model fails to generate meaningful objects, whereas inference-time scaling enables the production of plausible and semantically accurate content. 
Furthermore, in many cases, inference-time scaling improves alignment with fine-grained textual cues, leading to outputs more aligned with human preferences. 
For instance, prompts like ``\textit{A lamp casting a warm glow}'' and ``\textit{A lamp casting shadows on an old, forgotten map}'' are better grounded, faithfully reflecting elements such as warm illumination or the presence of a map surface.

\subsection{Additional Implementation Details}\label{app:implemtation}

\subsubsection{More Details of Base Models}\label{app:base_models}\hfill

\noindent
We investigate the implementation details of two base models, including GaussianCube~\cite{zhang2024gaussiancube} and TRELLIS~\cite{xiang2024structured}.

\noindent\textbf{Implementation details of GaussianCube.}
We adopt GaussianCube~\cite{zhang2024gaussiancube} as a baseline method due to its recent success in enabling structured and fully explicit radiance representations for 3D generative modeling. 
Unlike hybrid NeRF-based approaches that rely on implicit decoders—often leading to degraded expressiveness in generative settings—GaussianCube achieves high-fidelity and compact modeling by first applying a densification-constrained Gaussian fitting algorithm and then aligning the fitted Gaussians~\cite{kerbl3Dgaussians} into a predefined voxel grid via Optimal Transport. 
This structured representation allows the use of standard 3D convolutional architectures (\eg, 3D U-Net) without the need for custom network designs. 
Moreover, GaussianCube achieves comparable visual quality with one to two orders of magnitude fewer parameters than previous structured representations, significantly reducing the complexity of 3D diffusion-based generation. 
Its strong performance across various tasks, including unconditional object generation, class-conditioned synthesis, and text-to-3D modeling, demonstrates its effectiveness and justifies its selection as a strong baseline in our experiments.
To generate 3D objects from Gaussian noise, we follow the official implementation of GaussianCube\footnote{\url{https://github.com/GaussianCube/GaussianCube}}.

\textbf{Compressed search space.}
We define the search space for GaussianCube as the set of initial parameters from which the generative process begins. Specifically, this corresponds to the initialization tensor $\mathbf{x}_T \in \mathbb{R}^{N_\text{v} \times N_\text{v} \times N_\text{v} \times C}$, where $N_\text{v} = 32$ denotes the voxel grid resolution along each spatial axis, and $C = 14$ represents the number of channels used to encode each Gaussian (\textit{e.g.}, position, scale, rotation, opacity, and appearance). Thus, the full search space is defined over a tensor of shape $(14, 32, 32, 32)$, which serves as GaussianCube’s starting noise for reverse diffusion.
To enable more efficient exploration and manipulation of this high-dimensional space, we introduce a \textit{compressed search space} by reshaping $\mathbf{x}_T$ into a 2D matrix of shape $(C_\text{s}, N_\text{s}, N_\text{s})$, where $C_\text{s} = 7$ and $N_\text{s} = 256$. This flattening aggregates spatial and channel dimensions while preserving essential structure.

\noindent\textbf{Implementation details of TRELLIS.}
We include TRELLIS~\cite{xiang2024structured} as a baseline due to its state-of-the-art performance and versatility in 3D generation. TRELLIS introduces a novel Structured LATent (SLAT) representation, which enables unified decoding into multiple 3D output formats, including radiance fields, 3D Gaussians, and meshes. 
This flexibility is achieved by combining a sparse 3D grid structure with dense multiview features extracted from a powerful pretrained vision foundation model. The SLAT representation captures both geometry and appearance information by assigning local latent codes to active voxels intersecting the object’s surface, thus balancing spatial structure and rich visual detail.

The TRELLIS generation pipeline operates in two stages. It first predicts the sparse voxel structure, and then infers latent features for the non-empty voxels. A rectified flow transformer is used as the generative backbone, which is adapted to handle the sparsity of the SLAT space. This architecture allows TRELLIS to support conditioning on either text or images, and to produce high-quality results without requiring any explicit 3D fitting during training.
Given its strong performance across multiple modalities and its representation-agnostic design, TRELLIS serves as a compelling baseline in our evaluation.
To generate 3D objects from Gaussian noise, we follow the official implementation of TRELLIS\footnote{\url{https://github.com/microsoft/TRELLIS}}.

\textbf{Compressed search space.}
TRELLIS adopts a two-stage generation framework based on its Structured LATent (SLAT) representation. In the first stage, a voxel-wise sparse structure is generated via a diffusion model operating over a coarse 3D grid. In the second stage, features corresponding to the non-empty voxels are generated using a sparse 3D convolution-based diffusion model and subsequently decoded into renderable formats such as 3D Gaussians or meshes. Due to the significant variability in output dimensionality and data sparsity in the second stage, we define the search space solely over the first-stage voxel diffusion model.

Specifically, the first-stage diffusion model outputs a binary occupancy grid of shape $(8, 16, 16, 16)$, where $8$ denotes the feature dimension used during generation. This tensor is initialized as Gaussian noise and serves as the input to the reverse diffusion process for structure prediction. To facilitate efficient manipulation and latent exploration, we reshape this tensor into a 2D compressed representation of shape $(C_\text{s}, N_\text{s}, N_\text{s})$, where $C_\text{s} = 8$ and $N_\text{s} = 64$, thus forming our compressed search space.

To mitigate sampling instability introduced by the second-stage diffusion model, where different noise realizations could lead to inconsistent outputs even for identical voxel structures, we introduce a fixed noise sampling strategy. Specifically, we pre-sample a global Gaussian noise tensor of shape $(64, 64, 64, C_\text{flow})$, where $C_\text{flow}$ denotes the channel dimension used by the rectified flow model. During inference, once the first-stage voxel coordinates are determined, the corresponding feature noise vectors for the second-stage generation are consistently sampled via coordinate indexing from the pre-sampled global tensor. This ensures deterministic behavior of the second stage conditioned on the same structural output from the first stage, allowing us to isolate the effect of the first-stage structure in our search process.

\subsubsection{More Details of Search Algorithms}\label{app:algorithm_flow}\hfill

\noindent
We detail three search algorithms with pseudo-code, including random search, zero-order search, and heuristic search.

\noindent\textbf{Pseudo-code for random search.}
More detailed pseudo-codes for vanilla random search and improved random search are presented in Algorithm~\ref{alg:random_search} and Algorithm~\ref{alg:random_search_compressed}, respectively.

\begin{algorithm}[h]
\caption{Random Search}
\label{alg:random_search}
\begin{algorithmic}[1]
\Require Number of candidates $N$, evaluation function $\mathcal{F}(\cdot)$
\Ensure Best candidate $\mathbf{x}^*$

\State Initialize a set of $N$ random Gaussian noise samples $\{\mathbf{x}_i\}_{i=1}^{N}$
\State Generate candidates from noise samples
\State Evaluate candidates using $\mathcal{F}(\mathbf{x}_i)$
\State Select the best candidate: 
\[
\mathbf{x}^* = \arg\max_{\mathbf{x}_i} \mathcal{F}(\mathbf{x}_i)
\]
\State \Return $\mathbf{x}^*$
\end{algorithmic}
\end{algorithm}

\begin{algorithm}[h]
\caption{Random Search with Compressed Search Space}
\label{alg:random_search_compressed}
\begin{algorithmic}[1]
\Require Number of candidates $N$, evaluation function $\mathcal{F}(\cdot)$, permutation strength $\eta$
\Ensure Best candidate $\mathbf{x}^*$
\State Sample one Gaussian noise candidate $\mathbf{x}_0$
\State Perform SVD: $\mathbf{x}_0 = \mathbf{U}\mathbf{\Sigma}\mathbf{V}^\top$, $\sigma_\text{init}$ is the diagonal values of $\mathbf{\Sigma}$, $\mathbf{x}^*=\mathbf{x}_0$ \Comment{Noise decomposition}
\For{$i = 1$ to $N$}
    \State Sample one singular value $\sigma_i$ from $\mathcal{N}(\sigma_\text{init}, \eta\mathbf{I})$ \Comment{Search on singular values}
    \State Reconstruct: $\widetilde{\mathbf{x}}_i = \mathbf{U} \mathbf{\Sigma}_{\sigma_i} \mathbf{V}^\top$
    \If{$\mathcal{F}(\widetilde{\mathbf{x}}_i) > \mathcal{F}(\mathbf{x}^*)$} \Comment{Singular space reset}
        \State Update the best candidate: $\mathbf{x}^* = \widetilde{\mathbf{x}}_i$
        \State Reset the singular space: $\mathbf{x}^* = \mathbf{U}\mathbf{\Sigma}\mathbf{V}^\top$, $\sigma_\text{init}$ is the diagonal values of $\mathbf{\Sigma}$
    \EndIf
\EndFor
\State \Return $\mathbf{x}^*$
\end{algorithmic}
\end{algorithm}

\noindent\textbf{Pseudo-code for zero-order search.}
More detailed pseudo-codes for vanilla zero-order search and improved zero-order search are presented in Algorithm~\ref{alg:zero_order_search} and Algorithm~\ref{alg:zero_order_search_compressed}, respectively.

\begin{algorithm}[h]
\caption{Zero-Order Search}
\label{alg:zero_order_search}
\begin{algorithmic}[1]
\Require Number of iterations $K$, number of candidates $N$, search radius $\lambda$, evaluation function $\mathcal{F}(\cdot)$
\Ensure Best candidate $\mathbf{x}^*$
\State Initialize a pivot $\mathbf{x}_0$ from Gaussian noise
\For{$t = 0$ to $K-1$}
    \State Sample $N$ perturbed candidates $\{\mathbf{x}_t'\}$ within radius $\lambda$ around $\mathbf{x}_t$
    \State Evaluate candidates using $\mathcal{F}(\mathbf{x}_t')$
    \State Update pivot: 
    \[
    \mathbf{x}_{t+1} = \arg\max_{\mathbf{x}_t'} \mathcal{F}(\mathbf{x}_t')
    \]
\EndFor
\State \Return $\mathbf{x}_K$
\end{algorithmic}
\end{algorithm}

\begin{algorithm}[h]
\caption{Zero-Order Search with Gaussian Normalization and Compressed Search Space}
\label{alg:zero_order_search_compressed}
\begin{algorithmic}[1]
\Require Number of iterations $K$, number of candidates $N$, search radius $\zeta$, convergence threshold $\gamma$, evaluation function $\mathcal{F}(\cdot)$
\Ensure Best candidate $\mathbf{x}^*$
\State Initialize a pivot $\mathbf{x}_0$ from Gaussian noise
\State Perform SVD: $\mathbf{x}_0 = \mathbf{U}\mathbf{\Sigma}\mathbf{V}^\top$, $\sigma_0$ is the diagonal values of $\mathbf{\Sigma}$ \Comment{Noise decomposition}
\For{$t = 0$ to $K-1$}
    \State Sample $N$ perturbed candidates $\{\sigma'_t\}$ within radius $\lambda$ around $\sigma_t$ \Comment{Search on singular values}
    \State Reconstruct $N$ noise $\{\mathbf{x}_{\sigma'_t}\}$
    \State \textbf{Apply Gaussian normalization}: 
    \[
        \mathbf{x}_{\sigma'_t} = \frac{\mathbf{x}_{\sigma'_t} - \text{Mean}(\mathbf{x}_{\sigma'_t})}{\text{Var}(\mathbf{x}_{\sigma'_t})}
    \]
    \State Evaluate candidates: $\mathcal{S}_t = \mathcal{F}(\mathbf{x}_{\sigma'_t})$
    \State Update pivot: 
    \[
    \mathbf{\sigma}_{t+1} = \arg\max_{\mathbf{\sigma}'_t} \mathcal{F}(\mathbf{x}_{\sigma'_t})
    \]
    \If{ $\text{Var}(\mathcal{S}_t) < \zeta$} \Comment{Singular space reset}
        \State Reset the singular space: $\mathbf{x}_{\sigma_{t+1}} = \mathbf{U}\mathbf{\Sigma}\mathbf{V}^\top$, $\sigma_{t+1}$ is the diagonal values of $\mathbf{\Sigma}$
    \EndIf
\EndFor
\State\Return $\mathbf{x}_{\sigma_K}$
\end{algorithmic}
\end{algorithm}

\noindent\textbf{Pseudo-code for heuristic search.}
More detailed pseudo-codes for vanilla heuristic search and improved heuristic search are presented in Algorithm~\ref{alg:heuristic_search} and Algorithm~\ref{alg:heuristic_search_compressed}, respectively.

\begin{algorithm}[h]
\caption{Heuristic Search (Firefly Algorithm)}
\label{alg:heuristic_search}
\begin{algorithmic}[1]
\Require Number of iterations $K$, number of fireflies $N$, attractiveness factor $\beta_0$, light absorption coefficient $\gamma$, randomization parameter $\alpha$, evaluation function $\mathcal{F}(\cdot)$
\Ensure Best candidate $\mathbf{x}^*$

\State Initialize $N$ fireflies with random positions $\{\mathbf{x}_i\}_{i=1}^{N}$ from Gaussian noise
\For{$t = 0$ to $K-1$}
    \For{each firefly $i$}
        \For{each firefly $j \neq i$}
            \If{$\mathcal{F}(\mathbf{x}_j) > \mathcal{F}(\mathbf{x}_i)$}
                \State Compute Cartesian distance: $r_{ij} = \sqrt{\sum_{k=1}^d(\mathbf{x}_{i,k} - \mathbf{x}_{j,k})^2}$
                \State Compute attractiveness: $\beta = \beta_0 e^{-\gamma r_{ij}^2}$
                \State Update position: 
                \[
                \mathbf{x}_i = \mathbf{x}_i + \beta (\mathbf{x}_j - \mathbf{x}_i) + \alpha \bm{\epsilon}
                \]
            \EndIf
        \EndFor
    \EndFor
\EndFor
\State \Return $\mathbf{x}^* = \argmax_{\mathbf{x}_i} \mathcal{F}(\mathbf{x}_i)$
\end{algorithmic}
\end{algorithm}

\begin{algorithm}[h]
\caption{Heuristic Search (Firefly Algorithm) with Gaussian Normalization and Compressed Search Space}
\label{alg:heuristic_search_compressed}
\begin{algorithmic}[1]
\Require Number of iterations $K$, number of fireflies $N$, attractiveness factor $\beta_0$, light absorption coefficient $\gamma$, randomization parameter $\alpha$, evaluation function $\mathcal{F}(\cdot)$, permutation strength $\eta$
\Ensure Best candidate $\mathbf{x}^*$
\State Initialize a pivot $\mathbf{x}_\text{init}$ from Gaussian noise
\State Perform SVD: $\mathbf{x}_\text{init} = \mathbf{U}\mathbf{\Sigma}\mathbf{V}^\top$, $\sigma_\text{init}$ is the diagonal values of $\mathbf{\Sigma}$ \Comment{Noise decomposition}
\State Initialize $N$ fireflies$\{\mathbf{\sigma}_i\}_{i=1}^{N}$ from $\mathcal{N}(\sigma_\text{init}, \eta\mathbf{I})$ \Comment{Search on singular values}
\State Reconstruct $N$ noise $\{\mathbf{x}_{\sigma_i}\}$
\State \textbf{Apply Gaussian normalization}:
    \[
        \mathbf{x}_{\sigma_i} = \frac{\mathbf{x}_{\sigma_i} - \text{Mean}(\mathbf{x}_{\sigma_i})}{\text{Var}(\mathbf{x}_{\sigma_i})}
    \]
\State Evaluate brightness: $\mathcal{S} = \{\mathcal{F}(\mathbf{x}_{\sigma_i})\}^N_{i=1}$
\For{$t = 1$ to $K$}
    \For{each firefly $i$}
        \For{each firefly $j \neq i$}
            \If{$\mathcal{S}_j > \mathcal{S}_i$}
                \State Compute distance: $r_{ij} =\sqrt{\sum_{k=1}^d(\sigma_{i,k} - \sigma_{j,k})^2}$
                \State Compute attractiveness: $\beta = \beta_0 e^{-\gamma r_{ij}^2}$
                \State Update position:
                \[
                    \sigma_i = \sigma_i + \beta (\sigma_j - \mathbf{\sigma}_i) + \alpha \mathcal{N}(0, \mathbf{I})
                \]
                \State Update $\mathcal{S}_i$ with $\mathcal{F}(\mathbf{x}_{\sigma_i})$
            \EndIf
        \EndFor
    \EndFor
    \If{ $\text{Var}(\mathcal{S}) < \zeta$} \Comment{Singular space reset}
        \State Reset the singular space with best candidate
    \EndIf
\EndFor
\State\Return Best candidate $\mathbf{x}^* = \arg\max_{\mathbf{x}_i} \mathcal{S}$
\end{algorithmic}
\end{algorithm}

\subsection{Supplementary Experimental Settings}\label{app:add_settings}

\subsubsection{Details of Measurement Metrics}\label{app:metrics}\hfill

\noindent
In the main paper, we employ three evaluation strategies to demonstrate the superiority of the proposed method. Here we supplementary the details of the measurements.

\noindent\textbf{Evaluation with ImageReward.}
ImageReward~\cite{xu2024imagereward} is a text-to-image human preference reward model.
Due to its effectiveness, it has been broadly used for human preference evaluation in text-to-image generation~\cite{fan2024reinforcement} and text-to-3D generation~\cite{ye2025dreamreward}.
Given a (text, image) pair, it extracts image and text features, combines them with cross-attention, and uses an MLP to generate a scalar for preference comparison.
For each 3D asset, we uniformly render 120 RGB images from different viewpoints.
Afterward, the ImageReward score is computed from the multi-view renderings and averaged for each prompt.

\noindent\textbf{Evaluation with CLIPScore.}
The CLIP~\cite{radford2021clip} model is trained on a variety of (image, text) pairs. It can evaluate text-image alignment by computing the cosine distance between the image and text features. Following prior works~\cite{jain2021dreamfields,poole2022dreamfusion,wang2023prolificdreamer}, we calculate the CLIP similarity score for each 3D asset by rendering 120 views and averaging the scores across these views.

\noindent\textbf{Evaluation with GPTEval3D.}
We utilize GPTEval3D~\cite{wu2024gpt}, which is a comprehensive benchmark for text-to-3D generation evaluation. GPTEval3D includes 13 baseline methods $\mathcal{M}$, 110 text prompts, and 5 criteria that are text-asset alignment, 3D plausibility, texture details, geometry details, and texture-geometry coherency, respectively.
For a new method, GPTEval3D employs GPT-4V to compare 3D assets generated by this new method and one of the baseline methods with the same input text prompt.
These pairwise comparison results are then used to calculate the Elo rating for each model.
Specifically, let $\mathbf{A}$ be a matrix where $\mathrm{A}_{ij}$ represents the number of times that the $i$-th model outperforms the $j$-th model in comparisons. 
The Elo ratings for the models are computed by optimizing the following objective:
\begin{equation}
    \sigma = \arg\max\limits_{\sigma} \mathop{\sum}\limits_{i \neq j} \mathrm{A}_{ij} \log \left(1 + 10^{(\sigma_j - \sigma_i)/400}\right),
\end{equation}
where $\sigma_i \in \mathbb{R}$ is the Elo rating of the $i$-th model.
In this work, we calculate Elo ratings within the existing tournament, initializing and freezing baseline scores as specified in the official code\footnote{\href{https://github.com/3DTopia/GPTEval3D/blob/main/data/tournament-v0/config.json}{https://github.com/3DTopia/GPTEval3D/blob/main/data/tournament-v0/config.json}}. For interested readers, please kindly refer to~\cite{wu2024gpt}.

\subsection{Broader Impact Statement}
\label{sec:impact}

Our work explores inference-time scaling in text-to-3D generation by optimizing the initial Gaussian noise through a structured search framework. This approach enhances the quality of generated 3D assets without requiring additional training, which may lead to several positive societal impacts.

\noindent\textbf{Positive impacts.}  
Our method lowers the computational and technical barriers for high-quality 3D content creation. This has the potential to democratize 3D asset generation for artists, designers, and developers in industries such as gaming, film, AR/VR, and digital manufacturing. By avoiding the need for model retraining and enabling quality improvements purely at inference time, our framework can also contribute to more sustainable use of compute resources, extending the utility of existing 3D generative models.
Furthermore, by supporting high-quality generation from natural language prompts, our method can facilitate accessibility for users without 3D modeling expertise, including educators, creators, and researchers in fields like digital heritage, science communication, or robotics simulation. Furthermore, the proposed search mechanisms may inspire similar inference-time optimization strategies in other domains beyond 3D generation, such as video synthesis, robotics planning, or scientific simulations.

\noindent\textbf{Negative impacts.}  
We acknowledge several potential negative impacts. One concern is the misuse of high-fidelity 3D generation technologies in applications such as disinformation, deepfakes, or unauthorized content reproduction. Although our method does not introduce new capabilities in terms of realism compared to existing 3D diffusion models, it may lower the effort needed to generate visually convincing assets, which can exacerbate such risks.
Another possible issue is related to bias in text-to-3D generation. If the underlying generative models or vision-language encoders are trained on unbalanced or biased datasets, the improved generation process might reinforce or even amplify these biases. For instance, stereotypical object appearances or the exclusion of underrepresented categories could emerge in outputs.

\noindent\textbf{Mitigation and transparency.}  
To address these concerns, we advocate for the responsible release of the source code, including safety mechanisms such as watermarking and prompt filtering. 

\subsection{Reproducibility}
\label{sec:reproduce}

We provide implementation details, including illustrative algorithm descriptions and pseudo-code. The source code will be publicly released for reproducibility. 

\subsection{Limitations}
\label{sec:limitation}
In this work, we primarily focus on constructing an inference-time scaling framework for text-to-3D generation, enhancing search algorithms, and compressing the search space for improved efficiency and effectiveness. Currently, we rely on 2D/3D-aware verifiers. The integration of a semantic-aware verifier is important for fine-grained adjustments to 3D generation based on user inputs, but not currently included. For instance, users may want specific modifications such as ``\textit{I hope the ears of the tiger are longer.}''  Developing verifiers capable of interpreting such instructions and effectively guiding 3D generation will prompt more controllable and adaptable generation results. Therefore, we will prioritize the development and application of semantic-aware verifiers as the next research direction, enabling fine-grained and user-driven adjustments in text-to-3D generation.

\end{document}